\documentclass[letterpaper]{article}
\usepackage{aaai2026}
\nocopyright
\usepackage{times}
\usepackage{helvet}
\usepackage{courier}
\usepackage[hyphens]{url}
\usepackage{graphicx}
\usepackage{natbib}
\usepackage{amsmath}
\usepackage{amssymb}
\usepackage{amsfonts}
\usepackage{booktabs}
\usepackage{xcolor}
\usepackage{multirow}
\usepackage{placeins}
\usepackage{tikz}
\usetikzlibrary{arrows.meta,positioning,calc}

\graphicspath{{figures/}}
\setkeys{Gin}{draft=false}

\title{Three-Stage Learning Unlocks Strong Performance in Simple Models for Long-Term Time Series Forecasting}

\author{
    Zhenan Yu\textsuperscript{1},
    Guangxin Jiang\textsuperscript{1},
    Jin Yang\textsuperscript{1}\thanks{Corresponding author.}
}
\affiliations{
    \textsuperscript{1}Harbin Institute of Technology\\
    \{25S010055@stu.hit.edu.cn, gxjiang@hit.edu.cn, 22B910011@stu.hit.edu.cn\}
}

\begin{document}

\maketitle

\begin{abstract}
Recent studies on long-term time series forecasting have shown that simple linear models and MLP-based predictors can achieve strong performance without increasingly complex architectures. However, many competitive baselines still rely on structural priors such as frequency-domain modeling, explicit decomposition, multi-scale mixing, or sophisticated cross-variable interaction modules, while paying less attention to how simple temporal mappings should be trained and organized. In this paper, we propose STAIR, short for Stagewise Temporal Adaptation via Individualization and Residual Learning, a training paradigm for long-term time series forecasting that aims to unlock the capacity of simple temporal mapping models without introducing complex architectural modules. STAIR decomposes forecasting ability into three progressive stages: it first learns common temporal dynamics across variables through a shared temporal mapping, then adapts the shared model to each variable via channel-wise fine-tuning to capture variable-specific patterns, and finally complements the backbone with cross-variable information through residual learning. We further introduce Shared-to-Individual Fine-tuning and $\alpha$-RevIN to mitigate the limitations of strict channel independence and the overly strong normalization prior induced by standard RevIN. This design gradually increases modeling flexibility while keeping the core temporal predictor as a shallow MLP in the main experiments, with linear variants analyzed separately. Experiments on nine long-term forecasting benchmarks show that STAIR matches or outperforms recent strong baselines while preserving a simple temporal backbone, providing a concise and effective modeling perspective for long-term time series forecasting.
\end{abstract}

\section{Introduction}
Long-term time series forecasting (LTSF) is a fundamental task in time series analysis, with broad applications in energy management, traffic forecasting, weather modeling, and financial market analysis. Compared with short-term forecasting, long-term forecasting requires a model to maintain stable extrapolation and pattern recognition over a much longer prediction horizon. It has therefore often been associated with the need for stronger model expressiveness. Recent methods have improved forecasting performance by introducing Transformers, frequency-domain modeling, multi-scale mixing, explicit decomposition, and cross-variable interaction modules, achieving substantial progress on standard benchmarks~\cite{zhou2021informer,wu2021autoformer,zhou2022fedformer,wu2023timesnet,wang2024timemixer}.

However, recent evidence suggests that the difficulty of LTSF does not arise solely from architectural complexity. DLinear shows that simple linear models can match or even outperform complex Transformer-based models on several LTSF benchmarks~\cite{zeng2023dlinear}. Subsequent lightweight models such as FITS, SparseTSF, and TSMixer further demonstrate that linear mappings and shallow MLPs still possess strong modeling potential~\cite{xu2024fits,lin2024sparsetsf,chen2023tsmixer}. This trend motivates us to revisit a basic question: does improving long-term forecasting necessarily require increasingly complex architectural modules, or have the capabilities of simple models not yet been fully exploited?

We argue that a central challenge in LTSF lies not only in model complexity, but also in how different sources of forecasting information are trained and organized. These sources include temporal regularities shared across variables, variable-specific dynamics, and potential cross-variable dependencies. Existing channel-independent methods assume that different variables can share the same temporal forecasting logic, an assumption that has proven effective on many datasets~\cite{nie2023patchtst}. Nevertheless, this assumption may ignore variable-level heterogeneity and interactions. In contrast, training a fully separate model for each variable can capture variable-specific patterns, but may sacrifice shared statistical structure and increase parameter cost. Similarly, normalization methods such as RevIN alleviate non-stationarity~\cite{kim2022revin}, but standard RevIN imposes a strong prior by removing local mean and variance: it implicitly assumes that input-window statistics mostly reflect distribution shift rather than useful state information for prediction. As a result, standard RevIN is not uniformly suitable for all datasets.

Based on these observations, we propose STAIR, a training paradigm for LTSF named Stagewise Temporal Adaptation via Individualization and Residual Learning. Rather than training a complete model in a single end-to-end stage, STAIR decomposes forecasting ability into three progressive stages. The first stage learns common temporal dynamics through a shared temporal mapping across all variables. The second stage starts from the shared model and performs variable-wise fine-tuning to capture variable-specific patterns. The third stage further learns cross-variable residual information on top of the first two stages. With this stagewise training strategy, the model first learns stable shared regularities and then gradually releases variable-specific and cross-variable modeling capacity, thereby reducing the optimization difficulty of direct joint modeling.

We further introduce two simple improvements tailored to common LTSF modeling paradigms. First, we propose Shared-to-Individual Fine-tuning, which lies between channel-independent shared modeling and fully individual modeling. Instead of training a separate model for each variable from scratch, it initializes each variable-specific predictor from the shared temporal mapping, balancing shared temporal regularities and variable-level differences. Second, we propose $\alpha$-RevIN, which controls the degree of local statistic removal with a fixed strength. This creates a continuous transition between standard RevIN and no RevIN, mitigating the risk that standard RevIN removes useful mean or variance information too aggressively.

It is important to emphasize that STAIR does not rely on complex attention mechanisms, graph structures, frequency-domain filters, multi-scale mixing, or explicit temporal decomposition modules. The main backbone used in this paper is a shallow MLP temporal mapping, while linear variants are included as a capacity analysis. Our goal is not to propose a more complicated forecasting architecture, but to examine whether simple temporal mapping models can achieve competitive performance on standard long-term forecasting tasks when trained with an appropriate organization of inductive biases.

We evaluate STAIR on nine widely used LTSF benchmarks, including ETT, Electricity, Traffic, Weather, Exchange, and Solar. We follow the conventional LTSF setting with a fixed input length of 96 and prediction lengths of 96, 192, 336, and 720. Experimental results show that STAIR substantially improves the forecasting ability of a shallow MLP backbone under a unified setting, reaching competitive or even superior performance on multiple datasets compared with recent strong baselines. Further ablation studies verify the effects of shared temporal mapping, variable-wise fine-tuning, cross-variable residual learning, $\alpha$-RevIN, backbone capacity, and compatibility with existing forecasting models.

The main contributions of this paper are summarized as follows:
\begin{itemize}
    \item We propose STAIR, a stagewise training paradigm for long-term time series forecasting that progressively organizes forecasting ability into shared temporal regularities, variable-specific adaptation, and cross-variable residual learning.
    \item We introduce Shared-to-Individual Fine-tuning and $\alpha$-RevIN, providing a flexible variable-level adaptation scheme between channel-independent and fully individual modeling, while partially preserving local statistics to alleviate the overly strong prior of standard RevIN.
    \item We show that simple linear layers and shallow MLPs can remain highly competitive under a suitable training paradigm, offering a concise, interpretable, and efficient perspective for long-term time series forecasting.
\end{itemize}

\section{Related Work}
\subsection{Long-term Time Series Forecasting Models}

Long-term time series forecasting has received increasing attention in recent years. Early deep learning approaches mainly relied on recurrent networks, convolutional networks, or attention mechanisms to model long-range dependencies. With the success of Transformers in sequence modeling, methods such as Informer, Autoformer, and FEDformer were introduced into LTSF to address the efficiency and dependency modeling challenges of long sequences~\cite{zhou2021informer,wu2021autoformer,zhou2022fedformer}. Autoformer incorporates temporal decomposition as an internal module and uses an Auto-Correlation mechanism to model periodic dependencies. FEDformer further combines frequency-domain representations with seasonal-trend decomposition to improve global structure modeling. Later, PatchTST segments time series into patches and trains a Transformer in a channel-independent manner, achieving strong performance on multiple LTSF benchmarks~\cite{nie2023patchtst}. More recently, iTransformer reorganizes Transformer tokens along the variable dimension by treating each variable history as a token, thereby improving cross-variable modeling~\cite{liu2024itransformer}.

Beyond Transformers, several structured time series models have also been proposed. TimesNet transforms one-dimensional time series into two-dimensional tensors to capture multi-periodic variations~\cite{wu2023timesnet}. TimeMixer uses multi-scale decomposition and MLP mixing to capture temporal patterns at different sampling scales~\cite{wang2024timemixer}. ModernTCN revisits the potential of convolutional structures for time series tasks~\cite{luo2024moderntcn}. These methods indicate that effective long-term forecasting often depends on how temporal structure is organized, for example through decomposition, frequency-domain representations, multi-scale processing, patches, variable tokens, or convolutional receptive fields. In contrast, this paper does not introduce additional complex architectural modules. Instead, it studies how a stagewise training paradigm can organize shared regularities, variable-specific patterns, and cross-variable relationships in simple temporal mapping models.

\subsection{Linear and Lightweight MLP Models}

DLinear renewed interest in the effectiveness of simple models for LTSF by showing that a simple linear model can match or outperform complex Transformer-based methods on several standard benchmarks~\cite{zeng2023dlinear}. A series of lightweight methods further suggest that LTSF does not necessarily require highly complex architectures. For example, FITS performs interpolation in the frequency domain and achieves competitive performance with few parameters~\cite{xu2024fits}; SparseTSF reduces parameter cost through sparse cross-period prediction~\cite{lin2024sparsetsf}; and TSMixer uses pure MLP structures to mix information along temporal and variable dimensions, showing strong capability in multivariate forecasting~\cite{chen2023tsmixer}.

These works share with our paper the observation that simple models can be powerful, but our focus is different. Existing lightweight models mainly improve forecasting ability through architectural designs such as linear mappings, frequency-domain interpolation, sparse periodic modeling, or MLP mixing. We further argue that the performance of simple models depends not only on structural capacity, but also on how different levels of information are progressively released during training. STAIR uses a shallow MLP as the main backbone and studies linear variants in capacity analysis, while learning shared temporal regularities, variable-specific dynamics, and cross-variable residual information in separate stages.

\subsection{Channel Independence and Cross-variable Modeling}

Channel independence has become an important design paradigm in recent LTSF studies. PatchTST explicitly adopts a channel-independent setting, decomposing a multivariate series into multiple univariate series while sharing the same model parameters~\cite{nie2023patchtst}. This design reduces the complexity of multivariate joint modeling and avoids introducing excessive cross-variable parameters when variable dependencies are weak. The success of simple models such as DLinear also suggests that, for many LTSF datasets, univariate temporal patterns already contain a large amount of predictable information.

Nevertheless, channel independence imposes a strong prior: all variables share the same temporal forecasting logic, and cross-variable relationships are explicitly ignored in the main model. Fully individual modeling moves to the opposite extreme by training a separate model for each variable. Although it can capture variable-specific dynamics, it is less data-efficient and may lose shared statistical regularities. Meanwhile, works such as iTransformer and TSMixer show that cross-variable information can still be beneficial in some multivariate datasets~\cite{liu2024itransformer,chen2023tsmixer}. The proposed Shared-to-Individual Fine-tuning lies between channel-independent and individual modeling: the model first learns a temporal mapping shared by all variables, and then fine-tunes it for each variable. This preserves shared temporal regularities while allowing each variable to form its own adaptation.

\subsection{Normalization and Non-stationarity}

Non-stationarity is a major challenge in LTSF. RevIN addresses time-varying mean and variance by applying reversible instance normalization: it removes local statistics from each input instance and restores the corresponding statistics at the output~\cite{kim2022revin}. RevIN has been effective across multiple forecasting models and has become a common component in LTSF experiments.

Despite its effectiveness, RevIN also imposes a strong prior: local mean and variance are treated as statistical states to be removed before prediction and restored afterwards. In some datasets, however, changes in local mean or variance may themselves carry useful information about the future. Completely removing such information can harm prediction, especially when trends, scale changes, or distributional states persist into the future. Prior work has also noted that excessive stationarization may cause information loss. Non-stationary Transformer argues that over-stationarization can weaken a model's ability to distinguish genuine non-stationary events~\cite{liu2022nonstationary}. Dish-TS further points out that the input and prediction windows may follow different statistical distributions, so normalizing and denormalizing only with input-window statistics may be insufficient to characterize the output space~\cite{fan2023dishts}. Inspired by these observations, we propose $\alpha$-RevIN, which uses a fixed strength to control the degree of local statistic removal and compares mean-only, standard-deviation-only, and different partial-retention variants. The goal of $\alpha$-RevIN is not to replace RevIN, but to analyze and mitigate the limitations of the overly strong statistical prior induced by standard RevIN under a unified experimental setting.

\subsection{Position of This Work}

In summary, existing LTSF research has developed along two important directions. On one hand, complex models enhance expressiveness through decomposition, frequency-domain modeling, multi-scale processing, patches, or variable tokens. On the other hand, the success of linear and MLP-based models indicates that simple temporal mappings still have considerable potential. This paper follows the second direction, but focuses not only on whether the model structure is simple, but also on how simple models should be trained and organized. STAIR decomposes long-term forecasting ability into shared regularities, variable-specific adaptation, and cross-variable residuals, and models them progressively through stagewise training. This perspective is complementary to existing lightweight models and provides a unified way to understand channel independence, individual modeling, RevIN, and multivariate joint modeling.

\section{Method}
\begin{figure*}[t]
\centering
\includegraphics[width=0.95\textwidth,draft=false]{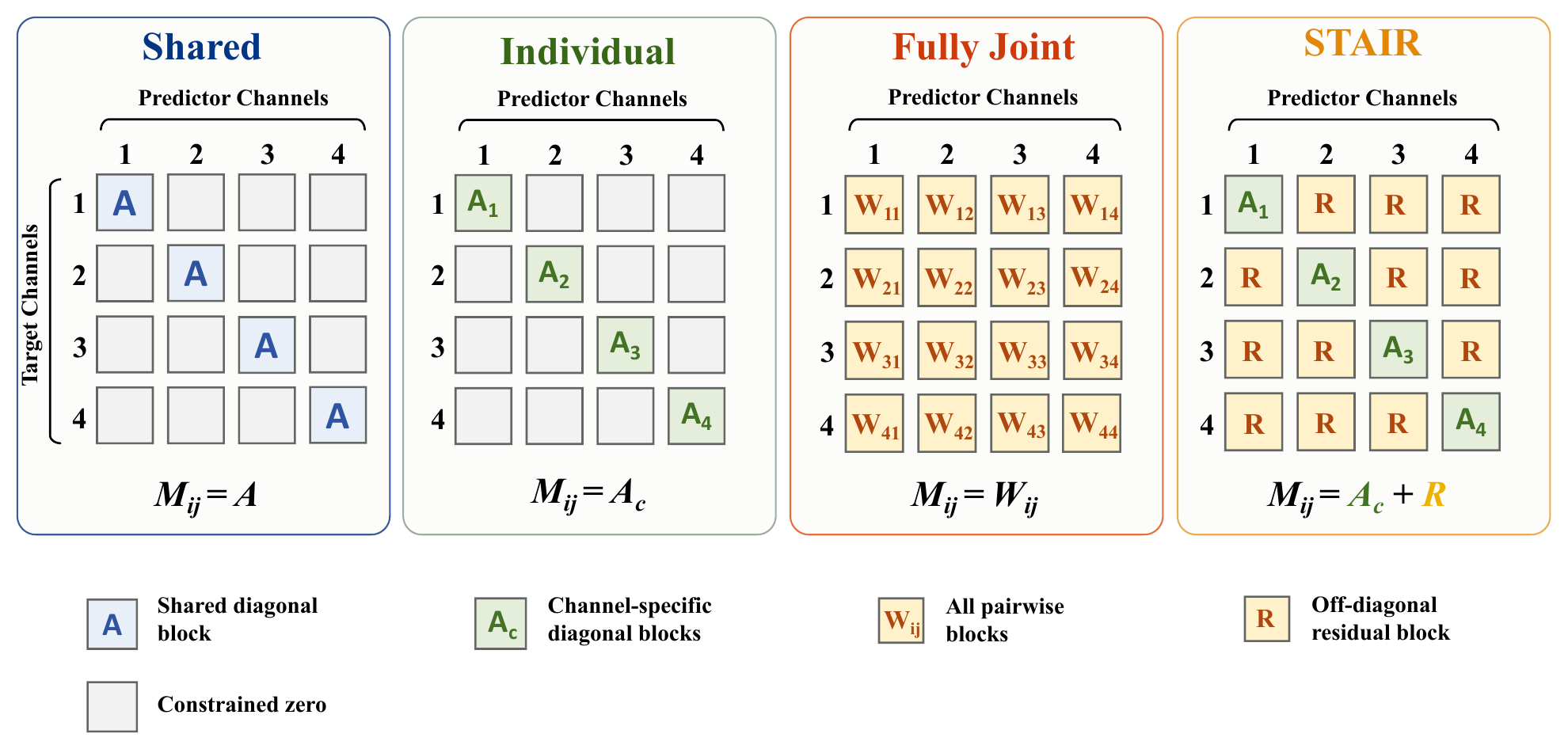}
\vspace{-0.4em}
\caption{A unified matrix view of channel modeling strategies. Channel-independent shared modeling uses identical diagonal temporal blocks, individual modeling relaxes them into channel-specific blocks, fully joint modeling learns all channel pairs, and STAIR adds controlled cross-variable residuals on top of an individual backbone.}
\label{fig:matrix_view}
\end{figure*}

\subsection{Problem Formulation}

Given a multivariate historical sequence of length $L$,
\[
\mathbf{X} = [\mathbf{x}_{1}, \mathbf{x}_{2}, \ldots, \mathbf{x}_{L}] \in \mathbb{R}^{L \times C},
\]
where $C$ denotes the number of variables, the goal of long-term time series forecasting is to predict a future sequence of length $H$:
\[
\mathbf{Y} = [\mathbf{x}_{L+1}, \mathbf{x}_{L+2}, \ldots, \mathbf{x}_{L+H}] \in \mathbb{R}^{H \times C}.
\]
For a mini-batch, the input and target are denoted as
\[
\mathbf{X} \in \mathbb{R}^{B \times L \times C}, \quad
\mathbf{Y} \in \mathbb{R}^{B \times H \times C}.
\]
This paper focuses on a simple temporal mapping model whose core objective is to learn the mapping from the historical window to the prediction window:
\[
f: \mathbb{R}^{L \times C} \rightarrow \mathbb{R}^{H \times C}.
\]
Unlike methods that introduce complex attention, frequency-domain, multi-scale, or decomposition modules, we use a linear layer or a shallow MLP as the basic temporal mapping and study how different levels of forecasting information can be organized through the training paradigm.

\subsection{A Unified View of Channel Modeling}

In multivariate LTSF, how to handle relationships among variables is a central issue. Existing approaches can be interpreted as imposing different constraints on the same joint mapping matrix. This matrix view is used as a structural lens rather than a restriction that the forecasting backbone must be linear. For nonlinear temporal mappings such as shallow MLPs, the same distinction corresponds to whether the temporal predictor is shared across variables, individualized for each variable, or complemented by explicit cross-variable pathways.

If the input sequence is flattened into a vector, a full multivariate linear mapping can be written as
\[
\mathrm{vec}(\hat{\mathbf{Y}}) = \mathbf{W}\mathrm{vec}(\mathbf{X}) + \mathbf{b},
\]
where
\[
\mathbf{W} \in \mathbb{R}^{HC \times LC}.
\]
This matrix can be viewed as a block matrix consisting of $C \times C$ blocks:
\[
\mathbf{W} =
\begin{bmatrix}
\mathbf{W}_{1,1} & \mathbf{W}_{1,2} & \cdots & \mathbf{W}_{1,C} \\
\mathbf{W}_{2,1} & \mathbf{W}_{2,2} & \cdots & \mathbf{W}_{2,C} \\
\vdots & \vdots & \ddots & \vdots \\
\mathbf{W}_{C,1} & \mathbf{W}_{C,2} & \cdots & \mathbf{W}_{C,C}
\end{bmatrix},
\]
where each block
\[
\mathbf{W}_{i,j} \in \mathbb{R}^{H \times L}
\]
represents the temporal mapping effect from the $j$-th input variable to the $i$-th output variable. From this perspective, three common channel modeling strategies can be understood in a unified manner.

\subsubsection{Fully Joint Modeling}

Fully joint multivariate modeling imposes no additional constraint on $\mathbf{W}$ and allows arbitrary mappings between variables:
\[
\mathbf{W}_{i,j} \neq 0, \quad \forall i,j.
\]
This strategy has the largest function space and theoretically contains channel-independent and individual modeling as special cases. However, its parameter count grows with $C^2$. When the number of variables is large, fully joint modeling may introduce substantial redundant parameters, increase optimization difficulty, and raise the risk of overfitting.

\subsubsection{Individual Modeling}

Individual modeling learns an independent temporal mapping for each variable, corresponding to a block-diagonal matrix:
\[
\mathbf{W}_{i,j}=0 \quad (i \neq j),
\]
with independent diagonal blocks:
\[
\mathbf{W}_{i,i} = \mathbf{A}_{i}.
\]
This strategy can capture variable-specific patterns, but it ignores temporal regularities shared across variables. Moreover, because each variable learns its own mapping independently, data efficiency may decrease when the number of training samples is limited.

\subsubsection{Channel-independent Shared Modeling}

Channel-independent shared modeling also removes all off-diagonal blocks, but all variables share the same temporal mapping:
\[
\mathbf{W}_{i,j}=0 \quad (i \neq j), \quad
\mathbf{W}_{i,i}=\mathbf{A}, \quad \forall i.
\]
This strategy has the smallest number of parameters and performs strongly on many LTSF benchmarks, suggesting that different variables may share common forecasting regularities. However, it also imposes a strong constraint: all variables must use exactly the same temporal mapping, and cross-variable effects are completely ignored. From the joint matrix perspective, channel-independent shared modeling sets all off-diagonal blocks to zero and forces all diagonal blocks to be identical. This constraint explains its parameter efficiency and stability, but also reveals its limitations: it cannot represent variable-level differences or explicitly exploit cross-variable effects.

These three strategies expose a fundamental tension. Fully joint modeling is expressive but difficult to optimize; channel-independent shared modeling is stable but overly constrained; individual modeling captures variable differences but is less data-efficient. We argue that a reasonable forecasting paradigm should not make a one-shot choice among these strategies. Instead, it should release model capacity progressively according to information reliability and optimization difficulty.

\begin{figure*}[t]
\centering
\includegraphics[width=0.95\textwidth,draft=false]{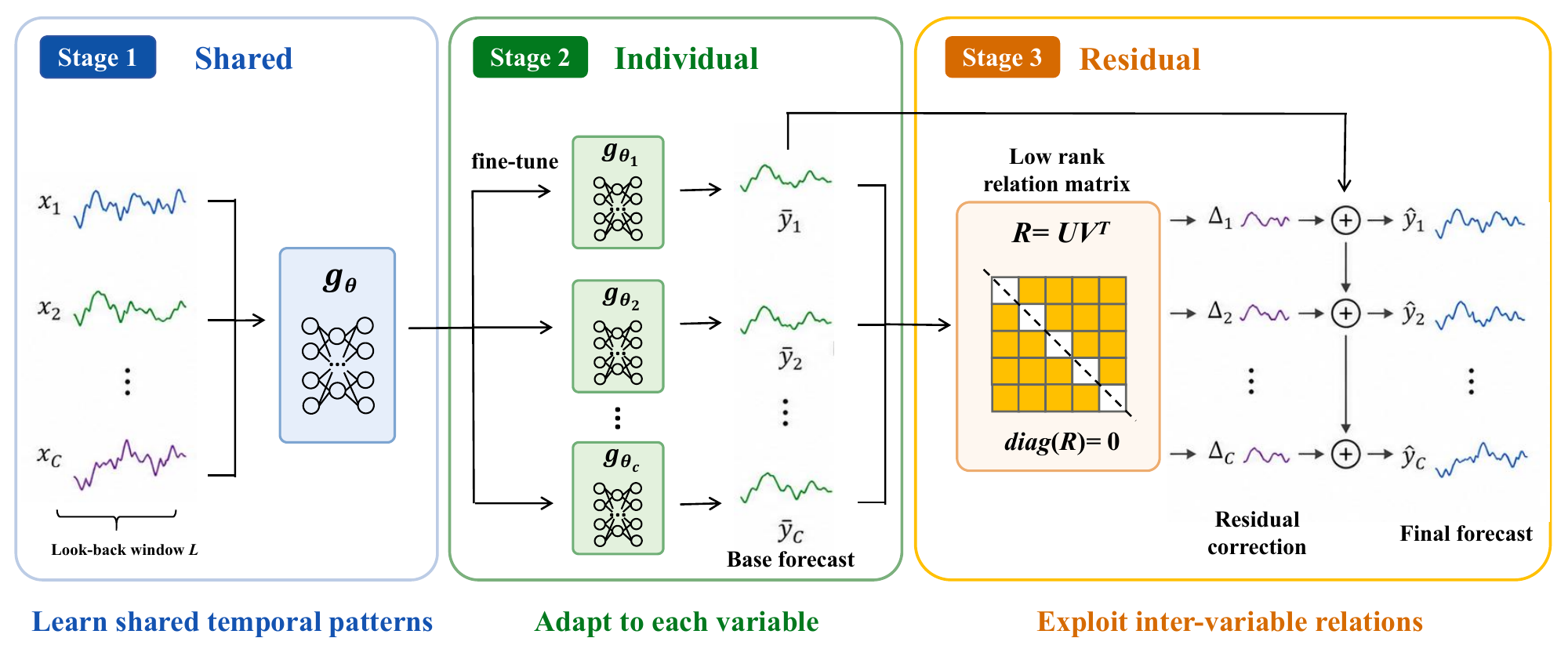}
\vspace{-0.4em}
\caption{Architecture overview of STAIR. A multivariate input window is first passed through a simple shared temporal mapping, which is then copied and fine-tuned into channel-specific predictors. The final stage freezes this backbone and adds a low-rank cross-variable residual correction.}
\label{fig:staged_framework}
\end{figure*}

\subsection{STAIR Framework}

Based on the unified view above, we propose STAIR, short for Stagewise Temporal Adaptation via Individualization and Residual Learning. The core idea is to divide forecasting information into three levels: shared temporal regularities, variable-specific patterns, and cross-variable residual information, and to model them sequentially through three training stages. We further hypothesize that these three types of information differ in reliability and optimization difficulty. Shared temporal regularities are supported by all variables and therefore have high data efficiency and stable generalization. Variable-specific patterns appear within individual variables and can complement local differences that the shared model cannot capture. Cross-variable dependencies may provide additional gains, but their effectiveness depends more strongly on dataset properties and can introduce extra parameters and overfitting risk. Therefore, STAIR does not release all modeling freedom at once, but follows the order of shared regularities, variable-specific adaptation, and cross-variable residual learning.

\subsubsection{Stage 1: Shared Temporal Mapping}

The first stage adopts channel-independent shared modeling. For each variable $c$, the model applies the same temporal mapping $g_{\theta}$:
\[
\hat{\mathbf{y}}^{(1)}_{c} = g_{\theta}(\mathbf{x}_{c}),
\]
where
\[
\mathbf{x}_{c} \in \mathbb{R}^{L}, \quad
\hat{\mathbf{y}}^{(1)}_{c} \in \mathbb{R}^{H}.
\]
The mapping $g_{\theta}$ can be either a linear layer or a shallow MLP. All variables share the parameters $\theta$, so this stage corresponds to
\[
\mathbf{W}_{i,i}=\mathbf{A}, \quad \mathbf{W}_{i,j}=0 \ (i \neq j).
\]

The objective of this stage is to learn temporal forecasting regularities that are common across variables. Since all variables jointly optimize the same mapping, the model can exploit shared statistical patterns in multivariate data while maintaining a small parameter count and stable optimization.

\subsubsection{Stage 2: Shared-to-Individual Fine-tuning}

The second stage performs variable-wise fine-tuning based on the Stage 1 model. Specifically, we initialize an independent mapping for each variable from the shared mapping $g_{\theta}$:
\[
g_{\theta} \rightarrow \{g_{\theta_{1}}, g_{\theta_{2}}, \ldots, g_{\theta_{C}}\}.
\]
The prediction for each variable is then generated by its corresponding mapping:
\[
\hat{\mathbf{y}}^{(2)}_{c} = g_{\theta_c}(\mathbf{x}_{c}).
\]
This stage corresponds to a block-diagonal matrix:
\[
\mathbf{W}_{i,i}=\mathbf{A}_{i}, \quad \mathbf{W}_{i,j}=0 \ (i \neq j),
\]
but differs from ordinary individual modeling because each $\mathbf{A}_{i}$ is initialized from the Stage 1 shared mapping $\mathbf{A}$ rather than random parameters.

This design allows the model to learn variable-specific patterns on top of shared regularities. In other words, Stage 2 does not abandon channel independence completely; instead, it treats channel independence as a shared prior and permits each variable to adapt around this prior. Compared with training individual models from scratch, this strategy is more data-efficient and more stable to optimize.

\subsubsection{Stage 3: Cross-variable Residual Learning}

The first two stages only model the temporal mapping of each variable itself and do not explicitly use cross-variable information. The third stage introduces cross-variable residual learning while freezing the Stage 2 backbone:
\[
\hat{\mathbf{Y}} = \hat{\mathbf{Y}}^{(2)} + r_{\phi}(\mathbf{X}),
\]
where $r_{\phi}$ complements the prediction with residual information induced by cross-variable dependencies. To avoid the overfitting risk caused by the $C^2$ parameter growth of fully joint modeling, Stage 3 only learns a lightweight residual module rather than retraining a full joint matrix.

In practice, this residual branch is implemented as a weak cross-variable adapter. Each variable history is first encoded into a compact representation, variables are then mixed by a low-rank relation matrix, and the mixed representation is decoded into a residual forecast:
\[
\mathbf{M}=\mathbf{U}\mathbf{V}^{\top}, \quad
\mathbf{R}=d_{\phi}\!\left(\mathbf{M}e_{\phi}(\mathbf{X})\right).
\]
The diagonal part of $\mathbf{M}$ is removed in the implementation so that the residual branch focuses on interactions between different variables rather than relearning the individual temporal backbone. This design is motivated by two practical considerations: cross-variable relations in many LTSF benchmarks are often weaker than each variable's own temporal dynamics, and in high-dimensional datasets such as Traffic, explicitly modeling all pairwise variable relations is computationally expensive and prone to overfitting. Therefore, STAIR captures cross-variable information only as a controlled low-rank residual correction.

From the joint matrix perspective, Stage 3 is equivalent to adding off-diagonal residuals on top of a learned block-diagonal matrix:
\[
\mathbf{W}
=
\begin{bmatrix}
\mathbf{A}_{1} & 0 & \cdots & 0 \\
0 & \mathbf{A}_{2} & \cdots & 0 \\
\vdots & \vdots & \ddots & \vdots \\
0 & 0 & \cdots & \mathbf{A}_{C}
\end{bmatrix}
+
\Delta \mathbf{W}_{\text{cross}}.
\]
Here, $\Delta \mathbf{W}_{\text{cross}}$ is responsible only for cross-variable residual information. This design follows our basic assumption: the dominant information in long-term forecasting comes from shared temporal regularities and variable-specific patterns, while cross-variable relationships usually serve as complementary information and should therefore be modeled as residuals with controlled parameter cost.

\subsection{$\alpha$-RevIN: Partial Reversible Normalization}

RevIN alleviates distribution shift in time series by removing local mean and variance for each input instance and restoring the corresponding statistics at the output. Given an input variable $\mathbf{x}_{c}$, standard RevIN can be written as
\[
\tilde{\mathbf{x}}_{c} = \frac{\mathbf{x}_{c}-\mu_c}{\sigma_c},
\]
followed by denormalization after prediction:
\[
\hat{\mathbf{y}}_{c} = \tilde{\mathbf{y}}_{c}\sigma_c+\mu_c.
\]

However, standard RevIN implies a strong assumption: local mean and variance changes are mainly non-stationary factors that should be removed. In some datasets, mean or scale changes may themselves contain useful information for future prediction. Standard RevIN also uses the input-window mean and variance for denormalization, which implies another assumption: the local level and scale of the prediction window are sufficiently close to those of the input window. When this assumption holds, RevIN can effectively reduce local distribution shift. When the prediction window undergoes substantial level switching or scale changes, however, restoring the output with input-window statistics may suppress true future changes. Moreover, even though denormalization adds the statistics back to the output, the main predictor can no longer directly use the instance-level statistical state that has been fully removed during modeling.

The original RevIN formulation often includes learnable affine parameters that scale and shift the normalized representation. However, these parameters do not directly control how much input-window statistics should be removed. In addition, the affine parameters are usually shared globally for each variable and apply the same scaling and shifting across different time windows, which may be insufficient for sequences with diverse window states. In this sense, the affine transform in standard RevIN acts more like a global correction after normalization than a decision about how much local statistical information should be preserved before normalization. In our experiments, relying only on affine parameters does not reliably resolve the issue of excessive statistic removal. Therefore, we do not use RevIN affine parameters and instead move the control mechanism into the normalization operation itself.

To mitigate these issues, we introduce $\alpha$-RevIN, which uses a fixed strength $\alpha$ to control the degree of local statistic removal:
\[
\tilde{\mathbf{x}}_{c}
=
\frac{\mathbf{x}_{c}-\alpha\mu_c}{\sigma_c^{\alpha}},
\]
and restores the prediction as
\[
\hat{\mathbf{y}}_{c}
=
\tilde{\mathbf{y}}_{c}\sigma_c^{\alpha}+\alpha\mu_c.
\]
When $\alpha=1$, $\alpha$-RevIN reduces to standard RevIN; when $\alpha=0$, it is equivalent to no RevIN; and when $0<\alpha<1$, the model partially removes local statistics, balancing normalization stability and statistical information preservation.

The key distinction is that $\alpha$-RevIN does not attempt to recover lost information only during denormalization. Instead, it avoids excessive information loss at the normalization stage. By preserving part of the instance-level level and scale information before the main predictor, the model can reduce local distribution shift while still exploiting statistical states that may be predictive. We also consider mean-only and standard-deviation-only variants to analyze the roles of level and scale information on different datasets. It should be emphasized that $\alpha$-RevIN does not change the final evaluation space: predictions are restored to the standard numerical scale before metrics are computed.

\subsection{Training Objective}

All stages use mean squared error as the main training objective:
\[
\mathcal{L}
=
\frac{1}{BHC}
\sum_{b=1}^{B}
\sum_{h=1}^{H}
\sum_{c=1}^{C}
\left(
\hat{Y}_{b,h,c}-Y_{b,h,c}
\right)^2.
\]
Stage 1, Stage 2, and Stage 3 are trained sequentially, and the best model in each stage is selected according to validation loss. Stage 1 learns shared temporal regularities, Stage 2 performs variable-wise fine-tuning initialized from the shared parameters, and Stage 3 learns cross-variable residuals while freezing the previous backbone. The final model is obtained by composing the three stages.

\section{Experiments}
This section systematically evaluates the effectiveness of the proposed method on long-term time series forecasting tasks. We mainly focus on the following questions: whether simple temporal mapping models can achieve competitive forecasting performance under the standard LTSF setting; how shared temporal regularities, variable-specific adaptation, and cross-variable residual information contribute in the three-stage training process; how $\alpha$-RevIN affects different datasets; and whether different datasets require different levels of model capacity.

\providecommand{\best}[1]{\textcolor{red}{\textbf{#1}}}
\providecommand{\second}[1]{\textcolor{blue}{\underline{#1}}}
\providecommand{\gain}[1]{#1\,{\raisebox{0.15ex}[0pt][0pt]{\scalebox{0.65}{\textcolor{blue}{$\downarrow$}}}}}

\begin{table}[!htbp]
\centering
\caption{Dataset statistics.}
\label{tab:datasets}
\resizebox{\linewidth}{!}{
\begin{tabular}{lcccl}
\toprule
Dataset & Variates & Frequency & Prediction Lengths & Train/Val/Test \\
\midrule
ETTh1 & 7 & 1 hour & \{96, 192, 336, 720\} & 8545 / 2881 / 2881 \\
ETTh2 & 7 & 1 hour & \{96, 192, 336, 720\} & 8545 / 2881 / 2881 \\
ETTm1 & 7 & 15 min & \{96, 192, 336, 720\} & 34465 / 11521 / 11521 \\
ETTm2 & 7 & 15 min & \{96, 192, 336, 720\} & 34465 / 11521 / 11521 \\
Electricity & 321 & 1 hour & \{96, 192, 336, 720\} & 18317 / 2633 / 5261 \\
Traffic & 862 & 1 hour & \{96, 192, 336, 720\} & 12185 / 1757 / 3509 \\
Weather & 21 & 10 min & \{96, 192, 336, 720\} & 36792 / 5271 / 10540 \\
Exchange & 8 & 1 day & \{96, 192, 336, 720\} & 5120 / 665 / 1422 \\
Solar & 137 & 10 min & \{96, 192, 336, 720\} & 36601 / 5161 / 10417 \\
\bottomrule
\end{tabular}
}
\end{table}

\begin{table*}[!t]
\centering
\caption{Long-term forecasting results averaged over four prediction lengths. The input length is fixed as 96 and the prediction lengths are $\{96,192,336,720\}$. Lower MSE/MAE indicates better forecasting performance. The best results are highlighted in \textcolor{red}{red bold}, and the second-best results are highlighted in \textcolor{blue}{blue underline}.}
\label{tab:main_results}
\scriptsize
\resizebox{\textwidth}{!}{
\begin{tabular}{l*{8}{cc}}
\toprule
Dataset & \multicolumn{2}{c}{\textbf{Ours}} & \multicolumn{2}{c}{\begin{tabular}{@{}c@{}}TimeMixer\\{\scriptsize 2024}\end{tabular}} & \multicolumn{2}{c}{\begin{tabular}{@{}c@{}}iTransformer\\{\scriptsize 2024}\end{tabular}} & \multicolumn{2}{c}{\begin{tabular}{@{}c@{}}PatchTST\\{\scriptsize 2023}\end{tabular}} & \multicolumn{2}{c}{\begin{tabular}{@{}c@{}}TimesNet\\{\scriptsize 2023}\end{tabular}} & \multicolumn{2}{c}{\begin{tabular}{@{}c@{}}Crossformer\\{\scriptsize 2023}\end{tabular}} & \multicolumn{2}{c}{\begin{tabular}{@{}c@{}}TiDE\\{\scriptsize 2023}\end{tabular}} & \multicolumn{2}{c}{\begin{tabular}{@{}c@{}}DLinear\\{\scriptsize 2023}\end{tabular}} \\
\cmidrule(lr){2-3}\cmidrule(lr){4-5}\cmidrule(lr){6-7}\cmidrule(lr){8-9}\cmidrule(lr){10-11}\cmidrule(lr){12-13}\cmidrule(lr){14-15}\cmidrule(lr){16-17}
Metric & MSE & MAE & MSE & MAE & MSE & MAE & MSE & MAE & MSE & MAE & MSE & MAE & MSE & MAE & MSE & MAE \\
\midrule
Weather & \second{0.246} & \second{0.273} & \best{0.240} & \best{0.272} & 0.258 & 0.278 & 0.259 & 0.280 & 0.259 & 0.286 & 0.259 & 0.315 & 0.270 & 0.320 & 0.265 & 0.317 \\
Solar & \best{0.215} & \second{0.269} & \second{0.216} & 0.280 & 0.233 & \best{0.262} & 0.270 & 0.307 & 0.301 & 0.319 & 0.641 & 0.639 & 0.347 & 0.417 & 0.330 & 0.401 \\
Electricity & \best{0.174} & \second{0.272} & 0.182 & 0.273 & \second{0.178} & \best{0.270} & 0.205 & 0.290 & 0.193 & 0.295 & 0.244 & 0.334 & 0.252 & 0.344 & 0.212 & 0.300 \\
Traffic & \second{0.466} & 0.303 & 0.485 & \second{0.297} & \best{0.428} & \best{0.282} & 0.481 & 0.304 & 0.620 & 0.336 & 0.550 & 0.304 & 0.760 & 0.473 & 0.625 & 0.383 \\
ETTh1 & \best{0.445} & \best{0.433} & \second{0.447} & \second{0.440} & 0.454 & 0.448 & 0.469 & 0.455 & 0.458 & 0.450 & 0.529 & 0.522 & 0.541 & 0.507 & 0.456 & 0.452 \\
ETTh2 & \second{0.372} & \second{0.403} & \best{0.365} & \best{0.395} & 0.383 & 0.406 & 0.387 & 0.407 & 0.414 & 0.427 & 0.942 & 0.683 & 0.611 & 0.550 & 0.559 & 0.515 \\
ETTm1 & \best{0.376} & \best{0.391} & \second{0.381} & \second{0.396} & 0.407 & 0.409 & 0.387 & 0.400 & 0.400 & 0.406 & 0.513 & 0.495 & 0.419 & 0.419 & 0.403 & 0.407 \\
ETTm2 & \best{0.271} & \best{0.316} & \second{0.275} & \second{0.323} & 0.288 & 0.332 & 0.281 & 0.326 & 0.291 & 0.333 & 0.757 & 0.611 & 0.358 & 0.404 & 0.350 & 0.401 \\
Exchange & \best{0.315} & \best{0.381} & -- & -- & 0.360 & \second{0.403} & 0.366 & 0.404 & 0.416 & 0.443 & 0.940 & 0.707 & 0.370 & 0.413 & \second{0.354} & 0.414 \\
\bottomrule
\end{tabular}
}
\end{table*}

\subsection{Experimental Setup}

\textbf{Datasets.}
We conduct experiments on nine widely used LTSF benchmarks, including ETTh1, ETTh2, ETTm1, ETTm2, Electricity, Traffic, Weather, Exchange, and Solar. These datasets cover real-world scenarios such as electricity load, traffic flow, weather dynamics, exchange rates, and energy production, with diverse numbers of variables, sampling frequencies, and non-stationarity patterns. Following the standard LTSF setting, we fix the input length to 96 and evaluate prediction lengths of 96, 192, 336, and 720. Dataset statistics are shown in Table~\ref{tab:datasets}.

\textbf{Metrics.}
Following common practice in LTSF, we use mean squared error (MSE) and mean absolute error (MAE) as the main evaluation metrics. All metrics are computed after predictions are restored to the standard evaluation scale, ensuring fair comparison with existing benchmarks.

\textbf{Baselines.}
We compare with seven representative strong baselines proposed in or after 2023: TimeMixer, iTransformer, PatchTST, TimesNet, Crossformer, TiDE, and DLinear~\cite{wang2024timemixer,liu2024itransformer,nie2023patchtst,wu2023timesnet,zhang2023crossformer,das2023tide,zeng2023dlinear}. These models represent different lines of research, including multi-scale mixing, inverted variable-wise Transformers, patch-based Transformers, temporal variation modeling, cross-variable dependency modeling, MLP encoder-decoder design, and linear forecasting. For fair comparison, baseline values are preferentially taken from original papers or official repositories that explicitly report results with input length 96. If a protocol-matched result is unavailable for a dataset, the corresponding entry is left blank.

\textbf{Implementation Details.}
The main experiments use a shallow MLP temporal mapping as the backbone, so that the proposed training paradigm is evaluated with a simple but moderately expressive predictor. Linear temporal mappings are studied separately in the capacity analysis. All experiments follow the standard LTSF setting with input length 96 and prediction lengths of 96, 192, 336, and 720. Models are trained with MSE loss, and the best checkpoint in each stage is selected according to validation loss. Because different datasets have different sample sizes, variable counts, and temporal complexity, we use a lightweight Optuna search before the final staged run to select only the Stage 1 MLP depth and hidden dimension on the validation set. This search is treated as a practical capacity-selection step rather than a contribution of the method. Stage 1 learns a temporal mapping shared by all variables, Stage 2 initializes from the shared model and performs variable-wise fine-tuning, and Stage 3 freezes the previous backbone and learns a cross-variable residual. The main experiments use $\alpha$-RevIN with a fixed strength of $\alpha=0.99$, and ablation studies further compare different normalization strengths. All main results are reported with a single random seed unless otherwise specified.

\subsection{Main Results}

Table~\ref{tab:main_results} reports averaged results over four prediction lengths. Complete horizon-wise results are provided in Table~\ref{tab:appendix_full_results}. The Ours column uses the best validation-selected stage for each prediction length, with no test-set information used for model selection, and therefore reflects the final stagewise model selection protocol.

The results show that STAIR is competitive with recent forecasting architectures while using a substantially simpler temporal mapping. It achieves the best average MSE on Solar, ETTh1, ETTm1, ETTm2, and Exchange, and remains second-best on several other datasets. The strongest external architectures still have advantages in some settings, especially Traffic and parts of Weather and ETTh2, indicating that specialized architectural priors remain useful. Nevertheless, the main observation is that a simple temporal mapping can approach or surpass recent strong baselines when shared regularities, variable-specific adaptation, and residual cross-variable information are organized in a staged manner. This supports the central claim that training organization, not only backbone sophistication, is a decisive factor in LTSF.

\subsection{Ablation Study}

\begin{figure*}[t]
\centering
\includegraphics[width=0.95\textwidth,draft=false]{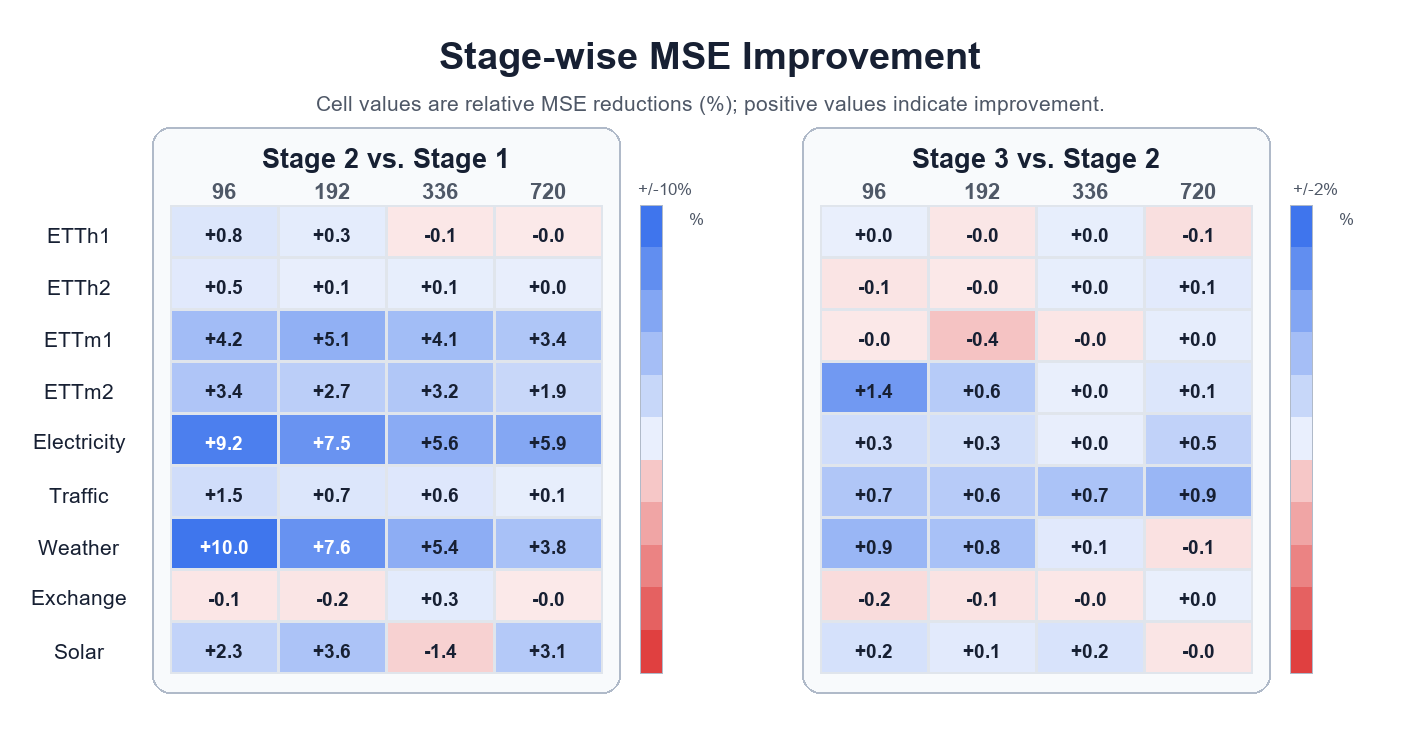}
\vspace{-0.4em}
\caption{Stage-wise MSE improvement heatmap. Each cell reports the relative MSE reduction from the previous stage to the current stage, where positive values indicate improvement. Stage 2 and Stage 3 use separate color scales because Stage 3 gains are generally smaller.}
\label{fig:stage_gain_heatmap}
\end{figure*}

\textbf{Effect of the Three Stages.}
Table~\ref{tab:stage_ablation} reports the averaged contribution of the three training stages. Stage 2 consistently reduces MSE on most datasets, indicating that although variables share certain temporal regularities, variable-level specificity remains important. Figure~\ref{fig:stage_gain_heatmap} further illustrates the relative gains at each prediction length. Stage 3 provides more moderate gains overall, but is more likely to bring additional improvements on high-dimensional datasets such as Traffic, Electricity, and Solar. This suggests that cross-variable residual information is more valuable in high-dimensional scenarios, while it tends to be weaker and less beneficial on low-dimensional datasets.

\begin{table}[!htbp]
\centering
\caption{Ablation study of the three-stage training framework. Results are averaged over four prediction lengths. Blue arrows indicate improvements over the previous stage.}
\label{tab:stage_ablation}
\scriptsize
\resizebox{\linewidth}{!}{
\begin{tabular}{lcccccc}
\toprule
\multirow{2}{*}{Dataset} & \multicolumn{2}{c}{Stage 1} & \multicolumn{2}{c}{Stage 2} & \multicolumn{2}{c}{Stage 3} \\
\cmidrule(lr){2-3}\cmidrule(lr){4-5}\cmidrule(lr){6-7}
& MSE & MAE & MSE & MAE & MSE & MAE \\
\midrule
Weather & 0.263 & 0.282 & \gain{0.246} & \gain{0.273} & \gain{0.246} & 0.273 \\
Solar & 0.221 & 0.272 & \gain{0.216} & 0.272 & \gain{0.216} & \gain{0.269} \\
Electricity & 0.187 & 0.280 & \gain{0.174} & \gain{0.272} & \gain{0.174} & \gain{0.271} \\
Traffic & 0.472 & 0.305 & \gain{0.469} & 0.308 & \gain{0.465} & \gain{0.303} \\
ETTh1 & 0.445 & 0.435 & \gain{0.444} & \gain{0.432} & 0.445 & \gain{0.432} \\
ETTh2 & 0.373 & 0.404 & \gain{0.372} & \gain{0.403} & \gain{0.372} & \gain{0.403} \\
ETTm1 & 0.393 & 0.401 & \gain{0.376} & \gain{0.391} & 0.377 & 0.391 \\
ETTm2 & 0.279 & 0.324 & \gain{0.271} & \gain{0.317} & \gain{0.270} & \gain{0.316} \\
Exchange & 0.315 & 0.381 & \gain{0.315} & \gain{0.381} & 0.315 & 0.381 \\
\bottomrule
\end{tabular}
}
\end{table}

\textbf{Observation.}
Table~\ref{tab:stage_ablation} shows that Stage 2 is the most reliable source of improvement: it reduces average MSE on almost all datasets and brings particularly clear gains on Weather, Electricity, ETTm1, and ETTm2. This indicates that the shared Stage 1 model learns useful common temporal structure, but the identical-mapping constraint is still too restrictive for many variables. Stage 3 brings smaller and more selective gains, especially on Traffic, Electricity, Solar, and ETTm2. This pattern is consistent with the design of STAIR: variable-specific adaptation should be introduced before cross-variable correction, while cross-variable information is best treated as a residual source rather than as the dominant modeling assumption.

\textbf{Compatibility with Existing Backbones.}
To examine whether the proposed stagewise training paradigm is tied to the shallow MLP backbone, we further apply the same three-stage logic to three representative forecasting backbones: DLinear, PatchTST, and TimeMixer-CI. DLinear is evaluated on all nine datasets. PatchTST and TimeMixer-CI are reported on the smaller datasets available in the current ablation runs, excluding the high-dimensional Traffic, Electricity, and Solar datasets. Tables~\ref{tab:dlinear_compatibility}--\ref{tab:timemixer_compatibility} report results averaged over the four prediction lengths. We only compute averages when all four horizons are available; incomplete or intentionally omitted runs are left blank. Full horizon-wise results are provided in the supplementary material.

\begin{table}[!htbp]
\centering
\caption{Compatibility with DLinear. Results are averaged over four prediction lengths. Blue arrows indicate improvements over the previous stage.}
\label{tab:dlinear_compatibility}
\scriptsize
\resizebox{\linewidth}{!}{
\begin{tabular}{lcccccc}
\toprule
\multirow{2}{*}{Dataset} & \multicolumn{2}{c}{Stage 1} & \multicolumn{2}{c}{Stage 2} & \multicolumn{2}{c}{Stage 3} \\
\cmidrule(lr){2-3}\cmidrule(lr){4-5}\cmidrule(lr){6-7}
& MSE & MAE & MSE & MAE & MSE & MAE \\
\midrule
ETTh1 & 0.456 & 0.453 & \gain{0.450} & \gain{0.438} & 0.479 & 0.460 \\
ETTh2 & 0.488 & 0.478 & 0.545 & 0.500 & 0.550 & 0.503 \\
ETTm1 & 0.404 & 0.411 & \gain{0.388} & \gain{0.395} & 0.411 & 0.413 \\
ETTm2 & 0.313 & 0.367 & 0.479 & 0.453 & \gain{0.477} & 0.454 \\
Electricity & 0.210 & 0.296 & \gain{0.205} & 0.297 & \gain{0.203} & 0.298 \\
Traffic & 0.626 & 0.385 & \gain{0.616} & 0.385 & \gain{0.597} & \gain{0.384} \\
Weather & 0.271 & 0.326 & \gain{0.247} & \gain{0.309} & \gain{0.246} & \gain{0.308} \\
Exchange & 0.294 & 0.372 & 0.295 & 0.373 & 0.295 & 0.377 \\
Solar & 0.327 & 0.400 & \gain{0.325} & \gain{0.399} & \gain{0.313} & \gain{0.376} \\
\bottomrule
\end{tabular}
}
\end{table}

\begin{table}[!htbp]
\centering
\caption{Compatibility with PatchTST. Results are averaged over four prediction lengths. Blank entries indicate unavailable complete-horizon runs.}
\label{tab:patchtst_compatibility}
\scriptsize
\resizebox{\linewidth}{!}{
\begin{tabular}{lcccccc}
\toprule
\multirow{2}{*}{Dataset} & \multicolumn{2}{c}{Stage 1} & \multicolumn{2}{c}{Stage 2} & \multicolumn{2}{c}{Stage 3} \\
\cmidrule(lr){2-3}\cmidrule(lr){4-5}\cmidrule(lr){6-7}
& MSE & MAE & MSE & MAE & MSE & MAE \\
\midrule
ETTh1 & 0.455 & 0.447 & 0.458 & \gain{0.447} & 0.477 & 0.458 \\
ETTh2 & 0.380 & 0.406 & \gain{0.378} & \gain{0.404} & 0.380 & 0.405 \\
ETTm1 & 0.385 & 0.401 & \gain{0.372} & \gain{0.393} & \gain{0.371} & 0.395 \\
ETTm2 & 0.287 & 0.332 & \gain{0.287} & 0.332 & 0.287 & 0.333 \\
Weather & 0.254 & 0.278 & \gain{0.251} & \gain{0.277} & \gain{0.244} & 0.286 \\
Exchange & 0.409 & 0.426 & \gain{0.399} & \gain{0.420} & \gain{0.398} & 0.420 \\
\bottomrule
\end{tabular}
}
\end{table}

\begin{table}[!htbp]
\centering
\caption{Compatibility with TimeMixer-CI. Results are averaged over four prediction lengths. Blank entries indicate unavailable complete-horizon runs.}
\label{tab:timemixer_compatibility}
\scriptsize
\resizebox{\linewidth}{!}{
\begin{tabular}{lcccccc}
\toprule
\multirow{2}{*}{Dataset} & \multicolumn{2}{c}{Stage 1} & \multicolumn{2}{c}{Stage 2} & \multicolumn{2}{c}{Stage 3} \\
\cmidrule(lr){2-3}\cmidrule(lr){4-5}\cmidrule(lr){6-7}
& MSE & MAE & MSE & MAE & MSE & MAE \\
\midrule
ETTh1 & 0.476 & 0.459 & \gain{0.465} & \gain{0.451} & 0.490 & 0.475 \\
ETTh2 & 0.385 & 0.409 & 0.390 & \gain{0.409} & \gain{0.390} & 0.409 \\
ETTm1 & 0.409 & 0.414 & \gain{0.408} & \gain{0.412} & 0.423 & 0.427 \\
ETTm2 & 0.282 & 0.327 & \gain{0.281} & \gain{0.325} & 0.282 & 0.328 \\
Weather & -- & -- & -- & -- & -- & -- \\
Exchange & 0.377 & 0.412 & \gain{0.375} & \gain{0.411} & \gain{0.374} & \gain{0.410} \\
\bottomrule
\end{tabular}
}
\end{table}

The results suggest that stagewise training is not restricted to the proposed MLP backbone. For DLinear, Stage 2 improves the averaged results on several datasets, including ETTh1, ETTm1, Electricity, Traffic, and Weather, showing that shared-to-individual adaptation is helpful even for a decomposed linear model. For PatchTST and TimeMixer-CI, the gains are more selective because their base representations are already stronger and more structured, but improvements still appear on multiple ETT, Weather, and Exchange settings. These results should be interpreted as compatibility evidence rather than exhaustive retuning of each backbone: the same training logic can provide additional gains on top of established architectures, while the magnitude depends on the inductive bias and capacity of the base model.

\textbf{Effect of $\alpha$-RevIN.}
We further analyze the effect of $\alpha$-RevIN. Standard RevIN assumes that local mean and variance mainly represent non-stationary factors to be removed, but this assumption does not hold uniformly across all datasets. Table~\ref{tab:revin_ablation} reports the normalization settings averaged over four prediction lengths.

\begin{table}[!htbp]
\centering
\caption{Ablation study of normalization strategies. Results are averaged over four prediction lengths. The best MSE and MAE in each row are highlighted in red.}
\label{tab:revin_ablation}
\scriptsize
\resizebox{\linewidth}{!}{
\begin{tabular}{lcccccccc}
\toprule
\multirow{2}{*}{Dataset} & \multicolumn{2}{c}{None} & \multicolumn{2}{c}{$\alpha=0.95$} & \multicolumn{2}{c}{$\alpha=0.99$} & \multicolumn{2}{c}{RevIN} \\
\cmidrule(lr){2-3}\cmidrule(lr){4-5}\cmidrule(lr){6-7}\cmidrule(lr){8-9}
& MSE & MAE & MSE & MAE & MSE & MAE & MSE & MAE \\
\midrule
ETTh1 & 0.457 & 0.453 & 0.458 & 0.449 & \best{0.445} & 0.435 & 0.447 & \best{0.433} \\
ETTh2 & 0.480 & 0.472 & 0.423 & 0.444 & \best{0.373} & 0.404 & 0.377 & \best{0.398} \\
ETTm1 & 0.403 & 0.410 & 0.405 & 0.408 & \best{0.401} & \best{0.402} & 0.412 & 0.405 \\
ETTm2 & 0.308 & 0.363 & 0.357 & 0.403 & \best{0.285} & 0.340 & 0.286 & \best{0.329} \\
Weather & 0.271 & 0.327 & 0.262 & 0.311 & \best{0.258} & 0.297 & 0.271 & \best{0.290} \\
Exchange & 0.298 & 0.378 & \best{0.246} & \best{0.358} & 0.315 & 0.381 & 0.355 & 0.398 \\
Traffic & 0.520 & 0.319 & 0.481 & 0.307 & \best{0.472} & \best{0.305} & 0.474 & 0.306 \\
Electricity & 0.193 & 0.287 & 0.189 & 0.282 & \best{0.187} & \best{0.280} & 0.194 & 0.281 \\
Solar & 0.227 & 0.274 & \best{0.223} & \best{0.273} & 0.229 & 0.276 & 0.256 & 0.281 \\
\bottomrule
\end{tabular}
}
\end{table}

Table~\ref{tab:revin_ablation} confirms that normalization is strongly dataset-dependent. On ETTh2, ETTm2, Weather, Traffic, and Electricity, removing most local statistics generally improves prediction, suggesting that local mean or scale variation often acts as distribution shift. In contrast, Exchange and Solar prefer a weaker $\alpha=0.95$ setting on average, indicating that their local level and scale still contain useful predictive state information. This explains why a fixed normalization choice is risky in LTSF. Local mean and variance are not always pure nuisance factors; depending on the dataset, they may represent distribution shift, predictive state, or both. The role of $\alpha$-RevIN is therefore to provide a controlled way to study and tune this statistical prior.

\textbf{Effect of Backbone Capacity.}
Finally, we compare a linear temporal mapping with the shallow MLP used in the main experiments. Linear layers are more interpretable and parameter-efficient, while shallow MLPs provide higher capacity and can capture certain nonlinear temporal patterns. This analysis is intended to clarify whether the gains of STAIR come only from increasing backbone capacity. The averaged results are shown in Table~\ref{tab:model_capacity}, and complete horizon-wise linear, MLP, and SOTA baseline comparisons are provided in the supplementary material.

\begin{table}[!htbp]
\centering
\caption{Effect of backbone capacity. Linear-STAIR and MLP-STAIR use fixed backbone families. The SOTA Baseline is the best published baseline value among the methods in Table~\ref{tab:main_results}.}
\label{tab:model_capacity}
\scriptsize
\resizebox{\linewidth}{!}{
\begin{tabular}{lcccccc}
\toprule
\multirow{2}{*}{Dataset} & \multicolumn{2}{c}{Linear-STAIR} & \multicolumn{2}{c}{MLP-STAIR} & \multicolumn{2}{c}{SOTA Baseline} \\
\cmidrule(lr){2-3}\cmidrule(lr){4-5}\cmidrule(lr){6-7}
& MSE & MAE & MSE & MAE & MSE & MAE \\
\midrule
Exchange & 0.315 & 0.381 & 0.368 & 0.408 & 0.350 & 0.398 \\
ETTh1 & 0.444 & 0.433 & 0.461 & 0.444 & 0.446 & 0.440 \\
ETTh2 & 0.372 & 0.403 & 0.380 & 0.405 & 0.365 & 0.395 \\
ETTm1 & 0.393 & 0.399 & 0.376 & 0.391 & 0.381 & 0.395 \\
ETTm2 & 0.281 & 0.336 & 0.270 & 0.316 & 0.275 & 0.323 \\
Weather & 0.236 & 0.276 & 0.246 & 0.273 & 0.238 & 0.272 \\
Traffic & -- & -- & 0.465 & 0.303 & 0.428 & 0.282 \\
Electricity & -- & -- & 0.174 & 0.271 & 0.177 & 0.268 \\
Solar & -- & -- & 0.215 & 0.269 & 0.216 & 0.262 \\
\bottomrule
\end{tabular}
}
\end{table}

The results indicate that the dominant regularities in several smaller datasets can already be captured by linear temporal mappings. Linear-STAIR is competitive on Exchange, ETTh1, ETTh2, and Weather, which suggests that these datasets contain strong low-complexity temporal structure under the staged training protocol. In contrast, ETTm1, ETTm2, Electricity, Traffic, and Solar benefit more from the nonlinear capacity of the shallow MLP. Therefore, increasing capacity is not the sole explanation for the gains of STAIR. The more important point is that model capacity should be released in a controlled order: first shared temporal regularities, then variable-specific adaptation, and finally residual cross-variable correction.

\FloatBarrier

\section{Conclusion}
This paper proposes STAIR, a Stagewise Temporal Adaptation via Individualization and Residual Learning paradigm for long-term time series forecasting. Instead of stacking increasingly complex architectural modules, we start from a unified matrix view of channel modeling and decompose effective forecasting information into shared temporal regularities, variable-specific dynamics, and cross-variable residual information. These components are then modeled progressively through three training stages. Stage 1 learns a temporal mapping shared by all variables, Stage 2 performs variable-wise fine-tuning initialized from the shared model, and Stage 3 further complements the backbone with cross-variable residual information. Through this stagewise training strategy, STAIR gradually releases the expressive capacity of simple temporal mapping models while preserving architectural simplicity.

We further introduce Shared-to-Individual Fine-tuning and $\alpha$-RevIN to mitigate the limitations of overly strong priors in channel independence and standard RevIN. Shared-to-Individual Fine-tuning lies between channel-independent shared modeling and fully individual modeling, preserving shared temporal regularities while allowing variable-level adaptation. $\alpha$-RevIN controls the degree of local statistic removal with a fixed strength, providing a simple way to analyze the roles of mean, variance, and local non-stationarity across different datasets.

Experiments on nine long-term forecasting benchmarks show that, when combined with the proposed stagewise training strategy, a shallow MLP backbone can achieve performance competitive with or superior to recent strong baselines on several datasets. Additional capacity analysis with linear temporal mappings further examines when simpler predictors are already sufficient. These results suggest that the core challenge of long-term time series forecasting does not arise solely from architectural complexity. How shared regularities, variable-specific information, and cross-variable relationships are trained and organized is also a crucial factor affecting forecasting performance.

Future work can extend this framework in two directions. First, more efficient and stable cross-variable residual modules may better adapt to high-dimensional multivariate datasets such as Traffic and Electricity. Second, the relationship between normalization strategies and dataset-specific non-stationarity deserves further study, so that useful instance-level statistics can be preserved while distribution shift is still mitigated under a unified experimental protocol.

\FloatBarrier

\bibliography{references}

\clearpage
\appendix
\section{Appendix Overview}

This appendix provides supplementary details beyond the main paper. Following common organization in LTSF papers, we include implementation details, dataset and preprocessing descriptions, complete result tables, additional ablation results, and a discussion of limitations. The main paper focuses on the core motivation, method, and main experimental conclusions; the appendix is mainly intended to improve reproducibility and provide space for complete experimental evidence.

\providecommand{\best}[1]{\textcolor{red}{\textbf{#1}}}
\providecommand{\second}[1]{\textcolor{blue}{\underline{#1}}}
\providecommand{\gain}[1]{#1\,{\raisebox{0.15ex}[0pt][0pt]{\scalebox{0.65}{\textcolor{blue}{$\downarrow$}}}}}

\section{Implementation Details}

\subsection{Overall Training Procedure}

The model is trained sequentially in three stages. In each stage, the best checkpoint is selected according to validation loss and used to initialize the next stage. The training objective is mean squared error (MSE), and all test metrics are computed after predictions are restored to the standard evaluation scale.

\begin{enumerate}
    \item \textbf{Stage 1: Shared temporal mapping.} We train a temporal mapping $g_{\theta}$ shared by all variables under the channel-independent setting.
    \item \textbf{Stage 2: Shared-to-individual fine-tuning.} We initialize the mapping for each variable from the Stage 1 shared mapping and perform variable-wise fine-tuning with a smaller learning rate. This stage may use anchor regularization to prevent the variable-specific models from drifting excessively away from the Stage 1 initialization.
    \item \textbf{Stage 3: Cross-variable residual learning.} We freeze the Stage 2 backbone and train only a lightweight low-rank cross-variable residual module. The residual decoder is zero-initialized so that the initial Stage 3 prediction is exactly equal to the Stage 2 prediction.
\end{enumerate}

\subsection{Temporal Mapping Architecture}

The temporal mapping can be either a one-layer linear layer or a shallow MLP operating along the temporal dimension. When the number of MLP layers is 1, the model reduces to a direct linear mapping from the look-back window to the prediction window. When multiple MLP layers are used, the hidden dimension and activation function need to be specified. If the activation function is \texttt{none}, dropout is skipped; otherwise, dropout is applied after hidden layers.

\begin{table}[h]
\centering
\caption{Default model and training hyperparameters used in the main experiments.}
\label{tab:appendix_hyperparams}
\resizebox{\linewidth}{!}{
\begin{tabular}{ll}
\toprule
Hyperparameter & Value \\
\midrule
Input length & 96 \\
Prediction lengths & \{96, 192, 336, 720\} \\
Batch size & 64 \\
Optimizer & Adam \\
Training loss & MSE \\
Stage 1 epochs & 20 \\
Stage 2 epochs & 20 \\
Stage 3 epochs & 20 \\
Stage 1 learning rate & $1\times 10^{-3}$ \\
Stage 2 learning rate & $1\times 10^{-5}$ \\
Stage 3 learning rate & $1\times 10^{-5}$ \\
Weight decay & $1\times 10^{-5}$ in each stage \\
Stage 2 anchor coefficient & $1\times 10^{-4}$ \\
Stage 3 hidden dimension & 32 \\
Stage 3 relation rank & 32 \\
Stage 3 residual scale & 1.0 \\
MLP hidden dimension & 512 \\
MLP activation & ReLU by default unless otherwise specified \\
Normalization & $\alpha$-RevIN by default unless otherwise specified \\
$\alpha$-RevIN strength & 0.99 in main experiments \\
RevIN affine parameters & Not used \\
Early stopping patience & 10 \\
Gradient clipping & 1.0 \\
Mixed precision training & Enabled on CUDA \\
Random seed & 2026 for main results \\
\bottomrule
\end{tabular}
}
\end{table}

The Stage 1 temporal mapping capacity is selected on the validation set before the three-stage run. We use Optuna with 15 trials and 4 startup trials; each trial trains the Stage 1 candidate for 20 epochs and selects the structure with the best validation MSE. The candidate layer numbers are $\{1,2,3,4\}$ and the candidate hidden dimensions are $\{64,96,128,192,256,512,1024\}$ for nonlinear MLPs; a one-layer setting corresponds to a linear temporal mapping. In the final main experiments, ETTh1, ETTh2, and Exchange use the linear temporal mapping. ETTm1, ETTm2, and Weather use a two-layer MLP with hidden dimension 512. Traffic, Electricity, and Solar use a four-layer MLP; Electricity uses hidden dimension 1024, while Traffic and Solar use hidden dimension 512.

\subsection{Low-rank Cross-variable Residual}

The Stage 3 residual module first encodes the historical sequence of each variable into a compact temporal representation, and then mixes information across variables through a low-rank variable relation matrix. The relation matrix is represented by two low-rank factors, which requires fewer parameters than a full $C \times C$ variable interaction matrix. In the implementation, the diagonal contribution of the low-rank relation matrix is removed so that the residual branch focuses on cross-variable correction rather than relearning the variable-wise backbone.

\subsection{Normalization Details}

For $\alpha$-RevIN, the model removes a fixed proportion of instance-level statistics before the predictor:
\[
\tilde{\mathbf{x}}_c = \frac{\mathbf{x}_c-\alpha\mu_c}{\sigma_c^\alpha}.
\]
After prediction, the output is restored to the standard evaluation scale:
\[
\hat{\mathbf{y}}_c = \tilde{\mathbf{y}}_c\sigma_c^\alpha + \alpha\mu_c.
\]
This paper does not use RevIN affine parameters. This design separates the control of normalization strength from the affine correction applied after normalization in standard RevIN.

\subsection{Metric Definitions}

For a test set with $N$ forecasting windows, prediction length $H$, and $C$ variables, MSE and MAE are computed as
\[
\mathrm{MSE} =
\frac{1}{NHC}\sum_{n=1}^{N}\sum_{h=1}^{H}\sum_{c=1}^{C}
\left(\hat{Y}_{n,h,c}-Y_{n,h,c}\right)^2,
\]
\[
\mathrm{MAE} =
\frac{1}{NHC}\sum_{n=1}^{N}\sum_{h=1}^{H}\sum_{c=1}^{C}
\left|\hat{Y}_{n,h,c}-Y_{n,h,c}\right|.
\]
All metrics are computed in the same evaluation space as the dataloader targets. When RevIN-family normalization is enabled, predictions are first denormalized by the corresponding input-window statistics before being compared with the target.

\subsection{Baseline Source Protocol}

The main comparison follows the common LTSF protocol with input length 96 and prediction lengths $\{96,192,336,720\}$. To avoid mixing incompatible settings, baseline results are copied from official paper tables only when the same input length and horizons are explicitly reported. The averaged main-table results are computed from the corresponding horizon-wise values. TimeMixer values are taken from the unified-hyperparameter long-term forecasting table in its appendix. The remaining reported baselines are taken from the full long-term forecasting table in the iTransformer appendix, which provides horizon-wise results for iTransformer, PatchTST, TimesNet, Crossformer, TiDE, and DLinear under the same input-length-96 setting.

\section{Datasets and Preprocessing Details}

\subsection{Datasets}

We conduct experiments on nine standard LTSF benchmarks, including ETTh1, ETTh2, ETTm1, ETTm2, Electricity, Traffic, Weather, Exchange, and Solar. Dataset statistics are shown in Table~\ref{tab:appendix_datasets}. We follow the standard LTSF setting with an input length of 96 and prediction lengths of \{96, 192, 336, 720\}.

\begin{table}[h]
\centering
\caption{Statistics of the datasets used in the experiments.}
\label{tab:appendix_datasets}
\resizebox{\linewidth}{!}{
\begin{tabular}{lcccl}
\toprule
Dataset & Variates & Frequency & Prediction Lengths & Train/Val/Test \\
\midrule
ETTh1 & 7 & 1 hour & \{96, 192, 336, 720\} & 8545 / 2881 / 2881 \\
ETTh2 & 7 & 1 hour & \{96, 192, 336, 720\} & 8545 / 2881 / 2881 \\
ETTm1 & 7 & 15 min & \{96, 192, 336, 720\} & 34465 / 11521 / 11521 \\
ETTm2 & 7 & 15 min & \{96, 192, 336, 720\} & 34465 / 11521 / 11521 \\
Electricity & 321 & 1 hour & \{96, 192, 336, 720\} & 18317 / 2633 / 5261 \\
Traffic & 862 & 1 hour & \{96, 192, 336, 720\} & 12185 / 1757 / 3509 \\
Weather & 21 & 10 min & \{96, 192, 336, 720\} & 36792 / 5271 / 10540 \\
Exchange & 8 & 1 day & \{96, 192, 336, 720\} & 5120 / 665 / 1422 \\
Solar & 137 & 10 min & \{96, 192, 336, 720\} & 36601 / 5161 / 10417 \\
\bottomrule
\end{tabular}
}
\end{table}

\subsection{Preprocessing}

Unless otherwise specified, all experiments use the multivariate-to-multivariate \texttt{M} setting. The dataset-level scaler is fitted only on the training split and then applied to the validation and test splits. If RevIN-family normalization is enabled, it is used as an internal window-level normalization operation and does not replace dataset-level standardization.

\subsection{Evaluation}

We report MSE and MAE. In each stage, the checkpoint with the lowest validation loss is selected for testing. Stage 1 and Stage 2 test results are used to analyze the contribution of different stages; unless otherwise specified by the experimental protocol, Stage 3 is treated as the final model output.

\section{Complete Main Forecasting Results}

This section reports the complete horizon-wise results of the final STAIR model under the selected Stage 1 capacity. The Ours column selects the best validated stage for each prediction length according to MSE, with MAE used as a secondary criterion when needed. Baseline values are transcribed from official paper tables under the input-length-96 LTSF protocol: TimeMixer is taken from Table 13 of \citet{wang2024timemixer}, and the remaining reported baselines are taken from Table 10 of \citet{liu2024itransformer}. TimeMixer does not report Exchange in that table, so the corresponding entries are left blank. We highlight the best and second-best values for each horizon and metric.

\providecommand{\best}[1]{\textcolor{red}{\textbf{#1}}}
\providecommand{\second}[1]{\textcolor{blue}{\underline{#1}}}
\providecommand{\gain}[1]{#1\,{\raisebox{0.15ex}[0pt][0pt]{\scalebox{0.65}{\textcolor{blue}{$\downarrow$}}}}}

The complete main results show that the selected-capacity STAIR model remains competitive across horizons, rather than only improving averaged scores. On ETTm and Weather, the gains are visible at most horizons, while on Exchange the linear selected structure is more appropriate than a larger MLP. On high-dimensional datasets such as Electricity and Traffic, the selected MLP capacity is needed to obtain competitive short- and medium-horizon performance. These horizon-wise results support the use of validation-based capacity selection before stagewise training.

\begin{table*}[p]
\centering
\scriptsize
\setlength{\tabcolsep}{2pt}
\renewcommand{\arraystretch}{1.08}
\caption{Complete horizon-wise main forecasting results. The best result is in red bold and the second-best result is in blue underline for each horizon and metric.}
\label{tab:appendix_full_results}
\resizebox{\textwidth}{!}{
\begin{tabular}{ll*{8}{cc}}
\toprule
\multirow{2}{*}{Dataset} & \multirow{2}{*}{Horizon} & \multicolumn{2}{c}{Ours} & \multicolumn{2}{c}{TimeMixer} & \multicolumn{2}{c}{iTransformer} & \multicolumn{2}{c}{PatchTST} & \multicolumn{2}{c}{TimesNet} & \multicolumn{2}{c}{Crossformer} & \multicolumn{2}{c}{TiDE} & \multicolumn{2}{c}{DLinear} \\
\cmidrule(lr){3-4}\cmidrule(lr){5-6}\cmidrule(lr){7-8}\cmidrule(lr){9-10}\cmidrule(lr){11-12}\cmidrule(lr){13-14}\cmidrule(lr){15-16}\cmidrule(lr){17-18}
& & MSE & MAE & MSE & MAE & MSE & MAE & MSE & MAE & MSE & MAE & MSE & MAE & MSE & MAE & MSE & MAE \\
\midrule
\multirow{4}{*}{ETTh1} & 96 & \second{0.382} & \best{0.389} & \best{0.375} & \second{0.400} & 0.386 & 0.405 & 0.414 & 0.419 & 0.384 & 0.402 & 0.423 & 0.448 & 0.479 & 0.464 & 0.386 & \second{0.400} \\
 & 192 & \second{0.433} & \best{0.420} & \best{0.429} & \second{0.421} & 0.441 & 0.436 & 0.460 & 0.445 & 0.436 & 0.429 & 0.471 & 0.474 & 0.525 & 0.492 & 0.437 & 0.432 \\
 & 336 & \best{0.477} & \best{0.446} & 0.484 & \second{0.458} & 0.487 & \second{0.458} & 0.501 & 0.466 & 0.491 & 0.469 & 0.570 & 0.546 & 0.565 & 0.515 & \second{0.481} & 0.459 \\
 & 720 & \best{0.486} & \best{0.479} & \second{0.498} & \second{0.482} & 0.503 & 0.491 & 0.500 & 0.488 & 0.521 & 0.500 & 0.653 & 0.621 & 0.594 & 0.558 & 0.519 & 0.516 \\
\midrule
\multirow{4}{*}{ETTh2} & 96 & \best{0.289} & \second{0.342} & \second{0.289} & \best{0.341} & 0.297 & 0.349 & 0.302 & 0.348 & 0.340 & 0.374 & 0.745 & 0.584 & 0.400 & 0.440 & 0.333 & 0.387 \\
 & 192 & \best{0.371} & \second{0.393} & \second{0.372} & \best{0.392} & 0.380 & 0.400 & 0.388 & 0.400 & 0.402 & 0.414 & 0.877 & 0.656 & 0.528 & 0.509 & 0.477 & 0.476 \\
 & 336 & \second{0.411} & \second{0.427} & \best{0.386} & \best{0.414} & 0.428 & 0.432 & 0.426 & 0.433 & 0.452 & 0.452 & 1.043 & 0.731 & 0.643 & 0.571 & 0.594 & 0.541 \\
 & 720 & \second{0.418} & 0.450 & \best{0.412} & \best{0.434} & 0.427 & \second{0.445} & 0.431 & 0.446 & 0.462 & 0.468 & 1.104 & 0.763 & 0.874 & 0.679 & 0.831 & 0.657 \\
\midrule
\multirow{4}{*}{ETTm1} & 96 & \best{0.313} & \best{0.354} & \second{0.320} & \second{0.357} & 0.334 & 0.368 & 0.329 & 0.367 & 0.338 & 0.375 & 0.404 & 0.426 & 0.364 & 0.387 & 0.345 & 0.372 \\
 & 192 & \best{0.354} & \best{0.376} & \second{0.361} & \second{0.381} & 0.377 & 0.391 & 0.367 & 0.385 & 0.374 & 0.387 & 0.450 & 0.451 & 0.398 & 0.404 & 0.380 & 0.389 \\
 & 336 & \best{0.386} & \best{0.398} & \second{0.390} & \second{0.404} & 0.426 & 0.420 & 0.399 & 0.410 & 0.410 & 0.411 & 0.532 & 0.515 & 0.428 & 0.425 & 0.413 & 0.413 \\
 & 720 & \best{0.452} & \best{0.435} & \second{0.454} & 0.441 & 0.491 & 0.459 & \second{0.454} & \second{0.439} & 0.478 & 0.450 & 0.666 & 0.589 & 0.487 & 0.461 & 0.474 & 0.453 \\
\midrule
\multirow{4}{*}{ETTm2} & 96 & \best{0.169} & \best{0.250} & \second{0.175} & \second{0.258} & 0.180 & 0.264 & \second{0.175} & 0.259 & 0.187 & 0.267 & 0.287 & 0.366 & 0.207 & 0.305 & 0.193 & 0.292 \\
 & 192 & \best{0.234} & \best{0.294} & \second{0.237} & \second{0.299} & 0.250 & 0.309 & 0.241 & 0.302 & 0.249 & 0.309 & 0.414 & 0.492 & 0.290 & 0.364 & 0.284 & 0.362 \\
 & 336 & \best{0.290} & \best{0.331} & \second{0.298} & \second{0.340} & 0.311 & 0.348 & 0.305 & 0.343 & 0.321 & 0.351 & 0.597 & 0.542 & 0.377 & 0.422 & 0.369 & 0.427 \\
 & 720 & \best{0.389} & \best{0.390} & \second{0.391} & \second{0.396} & 0.412 & 0.407 & 0.402 & 0.400 & 0.408 & 0.403 & 1.730 & 1.042 & 0.558 & 0.524 & 0.554 & 0.522 \\
\midrule
\multirow{4}{*}{Electricity} & 96 & \best{0.148} & \second{0.245} & \second{0.153} & 0.247 & \best{0.148} & \best{0.240} & 0.181 & 0.270 & 0.168 & 0.272 & 0.219 & 0.314 & 0.237 & 0.329 & 0.197 & 0.282 \\
 & 192 & \second{0.162} & 0.259 & 0.166 & \second{0.256} & \best{0.162} & \best{0.253} & 0.188 & 0.274 & 0.184 & 0.289 & 0.231 & 0.322 & 0.236 & 0.330 & 0.196 & 0.285 \\
 & 336 & \best{0.177} & \second{0.276} & 0.185 & 0.277 & \second{0.178} & \best{0.269} & 0.204 & 0.293 & 0.198 & 0.300 & 0.246 & 0.337 & 0.249 & 0.344 & 0.209 & 0.301 \\
 & 720 & \best{0.208} & \best{0.306} & 0.225 & \second{0.310} & 0.225 & 0.317 & 0.246 & 0.324 & \second{0.220} & 0.320 & 0.280 & 0.363 & 0.284 & 0.373 & 0.245 & 0.333 \\
\midrule
\multirow{4}{*}{Traffic} & 96 & \second{0.438} & 0.287 & 0.462 & \second{0.285} & \best{0.395} & \best{0.268} & 0.462 & 0.295 & 0.593 & 0.321 & 0.522 & 0.290 & 0.805 & 0.493 & 0.650 & 0.396 \\
 & 192 & \second{0.452} & 0.296 & 0.473 & 0.296 & \best{0.417} & \best{0.276} & 0.466 & 0.296 & 0.617 & 0.336 & 0.530 & \second{0.293} & 0.756 & 0.474 & 0.598 & 0.370 \\
 & 336 & \second{0.470} & 0.305 & 0.498 & \second{0.296} & \best{0.433} & \best{0.283} & 0.482 & 0.304 & 0.629 & 0.336 & 0.558 & 0.305 & 0.762 & 0.477 & 0.605 & 0.373 \\
 & 720 & \second{0.502} & 0.323 & 0.506 & \second{0.313} & \best{0.467} & \best{0.302} & 0.514 & 0.322 & 0.640 & 0.350 & 0.589 & 0.328 & 0.719 & 0.449 & 0.645 & 0.394 \\
\midrule
\multirow{4}{*}{Weather} & 96 & \second{0.163} & \best{0.208} & 0.163 & \second{0.209} & 0.174 & 0.214 & 0.177 & 0.218 & 0.172 & 0.220 & \best{0.158} & 0.230 & 0.202 & 0.261 & 0.196 & 0.255 \\
 & 192 & 0.209 & \best{0.249} & \second{0.208} & \second{0.250} & 0.221 & 0.254 & 0.225 & 0.259 & 0.219 & 0.261 & \best{0.206} & 0.277 & 0.242 & 0.298 & 0.237 & 0.296 \\
 & 336 & \second{0.266} & \second{0.292} & \best{0.251} & \best{0.287} & 0.278 & 0.296 & 0.278 & 0.297 & 0.280 & 0.306 & 0.272 & 0.335 & 0.287 & 0.335 & 0.283 & 0.335 \\
 & 720 & 0.346 & \second{0.343} & \best{0.339} & \best{0.341} & 0.358 & 0.347 & 0.354 & 0.348 & 0.365 & 0.359 & 0.398 & 0.418 & 0.351 & 0.386 & \second{0.345} & 0.381 \\
\midrule
\multirow{4}{*}{Exchange} & 96 & \best{0.080} & \best{0.201} & -- & -- & \second{0.086} & 0.206 & 0.088 & \second{0.205} & 0.107 & 0.234 & 0.256 & 0.367 & 0.094 & 0.218 & 0.088 & 0.218 \\
 & 192 & \best{0.161} & \best{0.289} & -- & -- & 0.177 & \second{0.299} & \second{0.176} & \second{0.299} & 0.226 & 0.344 & 0.470 & 0.509 & 0.184 & 0.307 & \second{0.176} & 0.315 \\
 & 336 & \best{0.300} & \second{0.398} & -- & -- & 0.331 & 0.417 & \second{0.301} & \best{0.397} & 0.367 & 0.448 & 1.268 & 0.883 & 0.349 & 0.431 & 0.313 & 0.427 \\
 & 720 & \best{0.719} & \best{0.636} & -- & -- & 0.847 & \second{0.691} & 0.901 & 0.714 & 0.964 & 0.746 & 1.767 & 1.068 & 0.852 & 0.698 & \second{0.839} & 0.695 \\
\midrule
\multirow{4}{*}{Solar} & 96 & \second{0.198} & \second{0.254} & \best{0.189} & 0.259 & 0.203 & \best{0.237} & 0.234 & 0.286 & 0.250 & 0.292 & 0.310 & 0.331 & 0.312 & 0.399 & 0.290 & 0.378 \\
 & 192 & \best{0.214} & \second{0.268} & \second{0.222} & 0.283 & 0.233 & \best{0.261} & 0.267 & 0.310 & 0.296 & 0.318 & 0.734 & 0.725 & 0.339 & 0.416 & 0.320 & 0.398 \\
 & 336 & \best{0.221} & \best{0.272} & \second{0.231} & 0.292 & 0.248 & \second{0.273} & 0.290 & 0.315 & 0.319 & 0.330 & 0.750 & 0.735 & 0.368 & 0.430 & 0.353 & 0.415 \\
 & 720 & \second{0.229} & \second{0.283} & \best{0.223} & 0.285 & 0.249 & \best{0.275} & 0.289 & 0.317 & 0.338 & 0.337 & 0.769 & 0.765 & 0.370 & 0.425 & 0.356 & 0.413 \\
\bottomrule
\end{tabular}
}
\end{table*}

\subsection{Complete Stage-wise Results}

Table~\ref{tab:appendix_stage_results} reports the full stage-wise results of the main STAIR model. Blue arrows mark improvements over the immediately previous stage under the same horizon and metric. The table shows that Stage 2 is the dominant source of improvement on many datasets, especially Weather, Electricity, ETTm1, and ETTm2. Stage 3 usually provides smaller but still useful corrections, with clearer gains on high-dimensional datasets such as Traffic, Electricity, and Solar. This pattern is consistent with the design motivation: variable-specific adaptation is more reliable and should be introduced before cross-variable residual correction.

\begin{table*}[p]
\centering
\tiny
\setlength{\tabcolsep}{3pt}
\renewcommand{\arraystretch}{0.58}
\caption{Complete stage-wise results of the final STAIR model. Blue arrows indicate improvements over the previous stage.}
\label{tab:appendix_stage_results}
\resizebox{\textwidth}{!}{
\begin{tabular}{llcccccc}
\toprule
\multirow{2}{*}{Dataset} & \multirow{2}{*}{Horizon} & \multicolumn{2}{c}{Stage 1} & \multicolumn{2}{c}{Stage 2} & \multicolumn{2}{c}{Stage 3} \\
\cmidrule(lr){3-4}\cmidrule(lr){5-6}\cmidrule(lr){7-8}
& & MSE & MAE & MSE & MAE & MSE & MAE \\
\midrule
\multirow{4}{*}{ETTh1} & 96 & 0.385 & 0.393 & \gain{0.382} & \gain{0.389} & \gain{0.382} & \gain{0.389} \\
 & 192 & 0.434 & 0.422 & \gain{0.433} & \gain{0.420} & 0.433 & 0.420 \\
 & 336 & 0.477 & 0.446 & 0.477 & \gain{0.443} & \gain{0.477} & \gain{0.443} \\
 & 720 & 0.486 & 0.479 & 0.486 & \gain{0.476} & 0.486 & \gain{0.475} \\
\midrule
\multirow{4}{*}{ETTh2} & 96 & 0.290 & 0.344 & \gain{0.289} & \gain{0.342} & 0.289 & \gain{0.342} \\
 & 192 & 0.371 & 0.393 & \gain{0.371} & \gain{0.393} & 0.371 & 0.393 \\
 & 336 & 0.411 & 0.427 & \gain{0.411} & \gain{0.427} & \gain{0.411} & 0.427 \\
 & 720 & 0.419 & 0.450 & \gain{0.418} & \gain{0.450} & \gain{0.418} & \gain{0.450} \\
\midrule
\multirow{4}{*}{ETTm1} & 96 & 0.327 & 0.365 & \gain{0.313} & \gain{0.354} & 0.313 & 0.354 \\
 & 192 & 0.374 & 0.388 & \gain{0.354} & \gain{0.376} & 0.356 & 0.377 \\
 & 336 & 0.402 & 0.408 & \gain{0.386} & \gain{0.398} & 0.386 & \gain{0.398} \\
 & 720 & 0.468 & 0.444 & \gain{0.452} & \gain{0.436} & \gain{0.452} & \gain{0.435} \\
\midrule
\multirow{4}{*}{ETTm2} & 96 & 0.177 & 0.260 & \gain{0.171} & \gain{0.252} & \gain{0.169} & \gain{0.250} \\
 & 192 & 0.242 & 0.301 & \gain{0.235} & \gain{0.295} & \gain{0.234} & \gain{0.294} \\
 & 336 & 0.300 & 0.340 & \gain{0.290} & \gain{0.331} & \gain{0.290} & \gain{0.331} \\
 & 720 & 0.397 & 0.397 & \gain{0.389} & \gain{0.391} & \gain{0.389} & \gain{0.390} \\
\midrule
\multirow{4}{*}{Exchange} & 96 & 0.080 & 0.201 & 0.080 & 0.201 & 0.080 & 0.202 \\
 & 192 & 0.161 & 0.289 & 0.161 & 0.289 & 0.161 & 0.289 \\
 & 336 & 0.301 & 0.399 & \gain{0.300} & \gain{0.398} & 0.300 & \gain{0.398} \\
 & 720 & 0.719 & 0.636 & 0.719 & \gain{0.636} & \gain{0.719} & \gain{0.636} \\
\midrule
\multirow{4}{*}{Weather} & 96 & 0.182 & 0.221 & \gain{0.164} & \gain{0.208} & \gain{0.163} & 0.208 \\
 & 192 & 0.228 & 0.260 & \gain{0.210} & \gain{0.250} & \gain{0.209} & \gain{0.249} \\
 & 336 & 0.281 & 0.297 & \gain{0.266} & \gain{0.291} & \gain{0.266} & 0.292 \\
 & 720 & 0.359 & 0.349 & \gain{0.346} & \gain{0.343} & 0.346 & 0.344 \\
\midrule
\multirow{4}{*}{Traffic} & 96 & 0.447 & 0.291 & \gain{0.441} & \gain{0.291} & \gain{0.438} & \gain{0.287} \\
 & 192 & 0.458 & 0.298 & \gain{0.455} & 0.301 & \gain{0.452} & \gain{0.296} \\
 & 336 & 0.476 & 0.307 & \gain{0.473} & 0.311 & \gain{0.470} & \gain{0.305} \\
 & 720 & 0.507 & 0.322 & \gain{0.506} & 0.331 & \gain{0.502} & \gain{0.323} \\
\midrule
\multirow{4}{*}{Electricity} & 96 & 0.163 & 0.255 & \gain{0.148} & \gain{0.246} & \gain{0.148} & \gain{0.245} \\
 & 192 & 0.176 & 0.268 & \gain{0.162} & \gain{0.259} & \gain{0.162} & \gain{0.259} \\
 & 336 & 0.187 & 0.282 & \gain{0.177} & \gain{0.276} & \gain{0.177} & \gain{0.276} \\
 & 720 & 0.222 & 0.315 & \gain{0.209} & \gain{0.307} & \gain{0.208} & \gain{0.306} \\
\midrule
\multirow{4}{*}{Solar} & 96 & 0.203 & 0.260 & \gain{0.198} & \gain{0.255} & \gain{0.198} & \gain{0.254} \\
 & 192 & 0.222 & 0.272 & \gain{0.214} & \gain{0.270} & \gain{0.214} & \gain{0.268} \\
 & 336 & 0.221 & 0.272 & 0.224 & 0.280 & \gain{0.224} & \gain{0.277} \\
 & 720 & 0.236 & 0.284 & \gain{0.229} & \gain{0.283} & 0.229 & \gain{0.279} \\
\bottomrule
\end{tabular}
}
\end{table*}

\section{Complete Backbone Compatibility Results}

This section reports the complete horizon-wise results for applying the three-stage training logic to existing forecasting backbones. Values are copied from the provided ablation CSV files and rounded to three decimals for presentation. Averages in the main paper are computed only when all four prediction lengths \{96, 192, 336, 720\} are available. Blue arrows indicate that the current stage improves over the immediately previous stage under the same dataset, horizon, and metric.

The backbone experiments are intended as compatibility evidence rather than a claim that each external architecture is fully retuned. DLinear, PatchTST, and TimeMixer-CI have different inductive biases, yet the full tables below show that stagewise adaptation often remains beneficial after replacing the simple MLP forecasting module. The effect is strongest when Stage 2 can correct variable-specific residuals without changing the backbone itself; Stage 3 is more dataset-dependent and is therefore reported horizon by horizon.

\begin{table*}[p]
\centering
\tiny
\setlength{\tabcolsep}{3pt}
\renewcommand{\arraystretch}{0.58}
\caption{Complete DLinear compatibility results. Blue arrows indicate improvements over the previous stage.}
\label{tab:appendix_dlinear_compatibility}
\resizebox{\textwidth}{!}{
\begin{tabular}{llcccccc}
\toprule
\multirow{2}{*}{Dataset} & \multirow{2}{*}{Horizon} & \multicolumn{2}{c}{Stage 1} & \multicolumn{2}{c}{Stage 2} & \multicolumn{2}{c}{Stage 3} \\
\cmidrule(lr){3-4}\cmidrule(lr){5-6}\cmidrule(lr){7-8}
& & MSE & MAE & MSE & MAE & MSE & MAE \\
\midrule
\multirow{4}{*}{ETTh1} & 96 & 0.387 & 0.404 & \gain{0.382} & \gain{0.394} & 0.382 & 0.394 \\
 & 192 & 0.441 & 0.438 & \gain{0.434} & \gain{0.426} & 0.434 & 0.426 \\
 & 336 & 0.482 & 0.460 & \gain{0.480} & \gain{0.446} & 0.600 & 0.532 \\
 & 720 & 0.515 & 0.510 & \gain{0.502} & \gain{0.486} & 0.502 & 0.487 \\
\midrule
\multirow{4}{*}{ETTh2} & 96 & 0.322 & 0.377 & 0.354 & 0.395 & 0.354 & 0.395 \\
 & 192 & 0.408 & 0.431 & 0.455 & 0.455 & 0.458 & 0.459 \\
 & 336 & 0.487 & 0.483 & 0.552 & 0.510 & 0.567 & 0.516 \\
 & 720 & 0.736 & 0.620 & 0.820 & 0.641 & 0.820 & 0.641 \\
\midrule
\multirow{4}{*}{ETTm1} & 96 & 0.342 & 0.373 & \gain{0.325} & \gain{0.358} & 0.326 & 0.359 \\
 & 192 & 0.385 & 0.398 & \gain{0.367} & \gain{0.381} & 0.379 & 0.394 \\
 & 336 & 0.414 & 0.415 & \gain{0.398} & \gain{0.402} & 0.406 & 0.408 \\
 & 720 & 0.477 & 0.457 & \gain{0.460} & \gain{0.438} & 0.531 & 0.492 \\
\midrule
\multirow{4}{*}{ETTm2} & 96 & 0.184 & 0.275 & 0.240 & 0.326 & \gain{0.236} & 0.328 \\
 & 192 & 0.248 & 0.320 & 0.362 & 0.401 & \gain{0.357} & 0.401 \\
 & 336 & 0.358 & 0.411 & 0.543 & 0.492 & 0.543 & 0.495 \\
 & 720 & 0.462 & 0.462 & 0.771 & 0.592 & 0.771 & 0.593 \\
\midrule
\multirow{4}{*}{Electricity} & 96 & 0.195 & 0.276 & \gain{0.187} & \gain{0.275} & \gain{0.182} & 0.277 \\
 & 192 & 0.195 & 0.282 & \gain{0.189} & \gain{0.281} & \gain{0.186} & 0.282 \\
 & 336 & 0.207 & 0.297 & \gain{0.204} & 0.299 & 0.204 & 0.299 \\
 & 720 & 0.243 & 0.329 & \gain{0.242} & 0.334 & 0.242 & 0.334 \\
\midrule
\multirow{4}{*}{Traffic} & 96 & 0.650 & 0.396 & \gain{0.636} & 0.397 & \gain{0.615} & 0.398 \\
 & 192 & 0.600 & 0.372 & \gain{0.589} & \gain{0.371} & \gain{0.576} & \gain{0.370} \\
 & 336 & 0.607 & 0.376 & \gain{0.598} & \gain{0.375} & \gain{0.581} & \gain{0.373} \\
 & 720 & 0.647 & 0.397 & \gain{0.640} & 0.398 & \gain{0.618} & \gain{0.393} \\
\midrule
\multirow{4}{*}{Weather} & 96 & 0.194 & 0.254 & \gain{0.167} & \gain{0.240} & \gain{0.166} & \gain{0.238} \\
 & 192 & 0.240 & 0.304 & \gain{0.213} & \gain{0.286} & 0.215 & 0.287 \\
 & 336 & 0.294 & 0.349 & \gain{0.266} & \gain{0.328} & \gain{0.265} & \gain{0.327} \\
 & 720 & 0.355 & 0.395 & \gain{0.342} & \gain{0.383} & \gain{0.338} & \gain{0.379} \\
\midrule
\multirow{4}{*}{Exchange} & 96 & 0.077 & 0.195 & 0.077 & 0.198 & 0.077 & \gain{0.196} \\
 & 192 & 0.153 & 0.284 & \gain{0.149} & \gain{0.283} & 0.168 & 0.300 \\
 & 336 & 0.268 & 0.386 & \gain{0.256} & \gain{0.380} & 0.301 & 0.416 \\
 & 720 & 0.679 & 0.624 & 0.697 & 0.633 & \gain{0.636} & \gain{0.594} \\
\midrule
\multirow{4}{*}{Solar} & 96 & 0.288 & 0.373 & \gain{0.286} & 0.374 & \gain{0.269} & \gain{0.350} \\
 & 192 & 0.316 & 0.397 & \gain{0.314} & \gain{0.394} & \gain{0.299} & \gain{0.369} \\
 & 336 & 0.349 & 0.414 & \gain{0.347} & \gain{0.412} & \gain{0.338} & \gain{0.392} \\
 & 720 & 0.354 & 0.414 & \gain{0.353} & 0.414 & \gain{0.345} & \gain{0.392} \\
\bottomrule
\end{tabular}
}
\end{table*}

The DLinear results provide a low-capacity reference. Improvements from Stage 2 are visible on several ETT, Weather, Traffic, and Electricity horizons, whereas very long ETTm2 horizons are less stable. This suggests that stagewise adaptation is helpful even for a decomposed linear backbone, but it cannot compensate for all long-horizon errors when the base decomposition is too restrictive.

\begin{table*}[p]
\centering
\scriptsize
\setlength{\tabcolsep}{3pt}
\renewcommand{\arraystretch}{0.86}
\caption{Complete PatchTST compatibility results. Blue arrows indicate improvements over the previous stage.}
\label{tab:appendix_patchtst_compatibility}
\resizebox{\textwidth}{!}{
\begin{tabular}{llcccccc}
\toprule
\multirow{2}{*}{Dataset} & \multirow{2}{*}{Horizon} & \multicolumn{2}{c}{Stage 1} & \multicolumn{2}{c}{Stage 2} & \multicolumn{2}{c}{Stage 3} \\
\cmidrule(lr){3-4}\cmidrule(lr){5-6}\cmidrule(lr){7-8}
& & MSE & MAE & MSE & MAE & MSE & MAE \\
\midrule
\multirow{4}{*}{ETTh1} & 96 & 0.405 & 0.412 & 0.405 & 0.412 & 0.407 & 0.413 \\
 & 192 & 0.439 & 0.431 & \gain{0.427} & \gain{0.424} & 0.431 & 0.427 \\
 & 336 & 0.483 & 0.465 & 0.499 & 0.467 & 0.499 & 0.467 \\
 & 720 & 0.493 & 0.480 & 0.499 & 0.483 & 0.572 & 0.524 \\
\midrule
\multirow{4}{*}{ETTh2} & 96 & 0.298 & 0.351 & 0.298 & \gain{0.350} & \gain{0.297} & \gain{0.349} \\
 & 192 & 0.379 & 0.394 & \gain{0.372} & \gain{0.390} & 0.372 & 0.390 \\
 & 336 & 0.405 & 0.425 & 0.405 & 0.425 & 0.405 & \gain{0.424} \\
 & 720 & 0.436 & 0.453 & 0.438 & \gain{0.451} & 0.444 & 0.458 \\
\midrule
\multirow{4}{*}{ETTm1} & 96 & 0.339 & 0.371 & \gain{0.318} & \gain{0.358} & \gain{0.317} & 0.358 \\
 & 192 & 0.367 & 0.387 & \gain{0.357} & \gain{0.380} & 0.357 & 0.383 \\
 & 336 & 0.405 & 0.413 & \gain{0.385} & \gain{0.402} & \gain{0.383} & 0.405 \\
 & 720 & 0.429 & 0.432 & \gain{0.428} & \gain{0.431} & \gain{0.427} & 0.432 \\
\midrule
\multirow{4}{*}{ETTm2} & 96 & 0.179 & 0.263 & 0.179 & \gain{0.261} & 0.179 & 0.263 \\
 & 192 & 0.248 & 0.309 & 0.248 & 0.309 & 0.249 & 0.309 \\
 & 336 & 0.315 & 0.351 & \gain{0.313} & \gain{0.349} & 0.315 & 0.353 \\
 & 720 & 0.408 & 0.405 & 0.409 & 0.409 & \gain{0.406} & \gain{0.407} \\
\midrule
\multirow{4}{*}{Weather} & 96 & 0.170 & 0.213 & \gain{0.164} & \gain{0.209} & 0.164 & 0.214 \\
 & 192 & 0.217 & 0.256 & \gain{0.211} & \gain{0.253} & 0.211 & 0.264 \\
 & 336 & 0.275 & 0.298 & \gain{0.269} & \gain{0.295} & \gain{0.264} & 0.310 \\
 & 720 & 0.353 & 0.347 & 0.360 & 0.352 & \gain{0.338} & 0.356 \\
\midrule
\multirow{4}{*}{Exchange} & 96 & 0.092 & 0.210 & \gain{0.088} & \gain{0.206} & 0.088 & 0.206 \\
 & 192 & 0.184 & 0.306 & 0.188 & 0.308 & 0.189 & 0.309 \\
 & 336 & 0.348 & 0.430 & \gain{0.347} & \gain{0.426} & \gain{0.346} & 0.426 \\
 & 720 & 1.011 & 0.757 & \gain{0.972} & \gain{0.742} & \gain{0.970} & \gain{0.741} \\
\bottomrule
\end{tabular}
}
\end{table*}

PatchTST already has a stronger temporal representation through patching and Transformer layers. The additional gains are therefore smaller and more horizon-specific, but Stage 2 still improves many ETT, Weather, and Exchange settings. This supports the claim that the proposed training framework can be layered onto stronger sequence encoders without changing their core architecture.

\begin{table*}[p]
\centering
\scriptsize
\setlength{\tabcolsep}{3pt}
\renewcommand{\arraystretch}{0.86}
\caption{Complete TimeMixer-CI compatibility results. Blue arrows indicate improvements over the previous stage.}
\label{tab:appendix_timemixer_compatibility}
\resizebox{\textwidth}{!}{
\begin{tabular}{llcccccc}
\toprule
\multirow{2}{*}{Dataset} & \multirow{2}{*}{Horizon} & \multicolumn{2}{c}{Stage 1} & \multicolumn{2}{c}{Stage 2} & \multicolumn{2}{c}{Stage 3} \\
\cmidrule(lr){3-4}\cmidrule(lr){5-6}\cmidrule(lr){7-8}
& & MSE & MAE & MSE & MAE & MSE & MAE \\
\midrule
\multirow{4}{*}{ETTh1} & 96 & 0.385 & 0.405 & \gain{0.383} & \gain{0.404} & 0.388 & 0.410 \\
 & 192 & 0.471 & 0.455 & \gain{0.469} & \gain{0.454} & 0.480 & 0.464 \\
 & 336 & 0.523 & 0.476 & \gain{0.493} & \gain{0.459} & 0.540 & 0.500 \\
 & 720 & 0.524 & 0.499 & \gain{0.514} & \gain{0.488} & 0.552 & 0.526 \\
\midrule
\multirow{4}{*}{ETTh2} & 96 & 0.298 & 0.348 & 0.299 & 0.348 & 0.299 & 0.348 \\
 & 192 & 0.383 & 0.404 & \gain{0.381} & \gain{0.401} & 0.381 & 0.402 \\
 & 336 & 0.434 & 0.443 & 0.453 & 0.443 & 0.453 & 0.443 \\
 & 720 & 0.425 & 0.443 & 0.426 & 0.444 & \gain{0.425} & \gain{0.443} \\
\midrule
\multirow{4}{*}{ETTm1} & 96 & 0.332 & 0.370 & \gain{0.331} & \gain{0.367} & \gain{0.330} & 0.367 \\
 & 192 & 0.373 & 0.392 & 0.374 & \gain{0.390} & \gain{0.368} & 0.390 \\
 & 336 & 0.426 & 0.423 & \gain{0.412} & \gain{0.416} & \gain{0.407} & 0.417 \\
 & 720 & 0.505 & 0.471 & 0.516 & 0.477 & 0.586 & 0.536 \\
\midrule
\multirow{4}{*}{ETTm2} & 96 & 0.176 & 0.259 & 0.178 & \gain{0.258} & 0.179 & 0.260 \\
 & 192 & 0.238 & 0.298 & \gain{0.237} & \gain{0.297} & 0.237 & 0.297 \\
 & 336 & 0.307 & 0.345 & 0.307 & \gain{0.342} & 0.309 & 0.348 \\
 & 720 & 0.409 & 0.406 & \gain{0.404} & \gain{0.402} & \gain{0.403} & 0.405 \\
\midrule
\multirow{4}{*}{Exchange} & 96 & 0.086 & 0.205 & 0.086 & 0.206 & 0.086 & 0.206 \\
 & 192 & 0.178 & 0.300 & 0.179 & 0.301 & \gain{0.178} & \gain{0.300} \\
 & 336 & 0.355 & 0.434 & \gain{0.345} & \gain{0.428} & \gain{0.345} & \gain{0.428} \\
 & 720 & 0.890 & 0.710 & \gain{0.888} & \gain{0.709} & \gain{0.887} & \gain{0.709} \\
\bottomrule
\end{tabular}
}
\end{table*}

TimeMixer-CI evaluates compatibility with a recent multi-scale backbone under a channel-independent setting. The results are mixed but informative: some horizons benefit from Stage 2 or Stage 3, while others degrade after additional adaptation. We therefore use these experiments as supporting evidence that the framework is not tied to the MLP module, and we leave exhaustive retuning of each external backbone to future work.

\section{Additional Ablation Studies}

This section provides the complete ablation evidence behind the concise tables in the main paper. Following the style of recent LTSF papers, we use the main paper for averaged summaries and report horizon-wise tables in the appendix. The horizon-wise tables reveal whether a design is consistently helpful or only improves specific prediction lengths.

\subsection{Normalization Ablation}

We compare several normalization strategies, including no instance normalization, standard RevIN, and $\alpha$-RevIN with different fixed strengths. This ablation examines whether local instance statistics should be fully removed or partially preserved. The results show that the preferred normalization strength is dataset-dependent. Stronger normalization is helpful on several ETT, Weather, Traffic, and Electricity settings, whereas Exchange and Solar are more sensitive to preserving absolute level information. Therefore, the normalization module should be viewed as a statistical prior rather than a universally beneficial preprocessing step.

\begin{table*}[p]
\centering
\tiny
\setlength{\tabcolsep}{2.5pt}
\renewcommand{\arraystretch}{0.58}
\caption{Complete horizon-wise normalization ablation. Best values among normalization settings are marked in red.}
\label{tab:appendix_revin_full}
\resizebox{\textwidth}{!}{
\begin{tabular}{llcccccccc}
\toprule
\multirow{2}{*}{Dataset} & \multirow{2}{*}{Horizon} & \multicolumn{2}{c}{None} & \multicolumn{2}{c}{$\alpha=0.95$} & \multicolumn{2}{c}{$\alpha=0.99$} & \multicolumn{2}{c}{RevIN} \\
\cmidrule(lr){3-4}\cmidrule(lr){5-6}\cmidrule(lr){7-8}\cmidrule(lr){9-10}
& & MSE & MAE & MSE & MAE & MSE & MAE & MSE & MAE \\
\midrule
\multirow{4}{*}{ETTh1} & 96 & 0.387 & 0.402 & 0.385 & 0.396 & \best{0.385} & \best{0.393} & 0.386 & 0.394 \\
 & 192 & 0.444 & 0.440 & 0.437 & 0.430 & \best{0.434} & \best{0.422} & 0.437 & 0.424 \\
 & 336 & 0.482 & 0.460 & 0.486 & 0.459 & \best{0.477} & \best{0.446} & 0.482 & 0.446 \\
 & 720 & 0.516 & 0.510 & 0.522 & 0.513 & 0.486 & 0.479 & \best{0.482} & \best{0.470} \\
\midrule
\multirow{4}{*}{ETTh2} & 96 & 0.320 & 0.375 & 0.322 & 0.379 & \best{0.290} & 0.344 & 0.291 & \best{0.337} \\
 & 192 & 0.405 & 0.429 & 0.404 & 0.425 & \best{0.371} & 0.393 & 0.375 & \best{0.389} \\
 & 336 & 0.488 & 0.484 & 0.449 & 0.457 & \best{0.411} & 0.427 & 0.420 & \best{0.426} \\
 & 720 & 0.707 & 0.600 & 0.519 & 0.515 & \best{0.419} & 0.450 & 0.422 & \best{0.440} \\
\midrule
\multirow{4}{*}{ETTm1} & 96 & \best{0.342} & 0.372 & 0.344 & 0.371 & 0.344 & \best{0.369} & 0.349 & 0.370 \\
 & 192 & \best{0.381} & 0.393 & 0.383 & 0.391 & 0.383 & \best{0.389} & 0.389 & 0.389 \\
 & 336 & 0.414 & 0.417 & 0.415 & 0.414 & \best{0.404} & \best{0.406} & 0.422 & 0.411 \\
 & 720 & 0.476 & 0.456 & 0.479 & 0.456 & \best{0.475} & \best{0.445} & 0.487 & 0.449 \\
\midrule
\multirow{4}{*}{ETTm2} & 96 & 0.182 & \best{0.271} & 0.197 & 0.294 & 0.184 & 0.277 & \best{0.181} & 0.271 \\
 & 192 & 0.249 & 0.321 & 0.292 & 0.367 & 0.248 & 0.318 & \best{0.248} & \best{0.305} \\
 & 336 & 0.333 & 0.391 & 0.392 & 0.432 & \best{0.308} & 0.358 & 0.309 & \best{0.343} \\
 & 720 & 0.467 & 0.467 & 0.546 & 0.518 & \best{0.401} & 0.408 & 0.407 & \best{0.398} \\
\midrule
\multirow{4}{*}{Weather} & 96 & 0.199 & 0.264 & 0.192 & 0.250 & \best{0.192} & 0.243 & 0.192 & \best{0.233} \\
 & 192 & 0.239 & 0.304 & 0.233 & 0.290 & \best{0.229} & 0.279 & 0.238 & \best{0.270} \\
 & 336 & 0.292 & 0.347 & 0.279 & 0.328 & \best{0.274} & 0.313 & 0.291 & \best{0.306} \\
 & 720 & 0.353 & 0.393 & 0.342 & 0.377 & \best{0.338} & 0.355 & 0.364 & \best{0.354} \\
\midrule
\multirow{4}{*}{Exchange} & 96 & 0.080 & \best{0.199} & 0.080 & 0.204 & \best{0.080} & 0.201 & 0.083 & 0.199 \\
 & 192 & \best{0.153} & \best{0.287} & 0.165 & 0.301 & 0.161 & 0.289 & 0.173 & 0.294 \\
 & 336 & 0.270 & 0.393 & \best{0.261} & \best{0.391} & 0.301 & 0.399 & 0.320 & 0.408 \\
 & 720 & 0.690 & 0.632 & \best{0.478} & \best{0.535} & 0.719 & 0.636 & 0.843 & 0.691 \\
\midrule
\multirow{4}{*}{Traffic} & 96 & 0.494 & 0.307 & 0.458 & 0.296 & \best{0.447} & \best{0.291} & 0.448 & 0.292 \\
 & 192 & 0.503 & 0.311 & 0.464 & \best{0.297} & \best{0.458} & 0.298 & 0.459 & 0.299 \\
 & 336 & 0.518 & 0.327 & 0.485 & 0.309 & \best{0.476} & \best{0.307} & 0.477 & 0.307 \\
 & 720 & 0.567 & 0.334 & 0.517 & 0.328 & \best{0.507} & \best{0.322} & 0.514 & 0.328 \\
\midrule
\multirow{4}{*}{Electricity} & 96 & 0.169 & 0.264 & 0.165 & 0.257 & \best{0.163} & \best{0.255} & 0.167 & 0.256 \\
 & 192 & 0.183 & 0.276 & \best{0.175} & 0.268 & 0.176 & 0.268 & 0.178 & \best{0.267} \\
 & 336 & 0.194 & 0.289 & 0.190 & 0.285 & \best{0.187} & \best{0.282} & 0.194 & 0.282 \\
 & 720 & 0.227 & 0.318 & 0.225 & 0.320 & \best{0.222} & \best{0.315} & 0.237 & 0.319 \\
\midrule
\multirow{4}{*}{Solar} & 96 & 0.223 & 0.269 & \best{0.211} & 0.261 & 0.212 & \best{0.260} & 0.227 & 0.262 \\
 & 192 & \best{0.220} & \best{0.269} & 0.220 & 0.271 & 0.235 & 0.280 & 0.249 & 0.277 \\
 & 336 & \best{0.231} & \best{0.280} & 0.235 & 0.282 & 0.233 & 0.282 & 0.272 & 0.291 \\
 & 720 & 0.234 & 0.278 & \best{0.226} & \best{0.278} & 0.234 & 0.281 & 0.277 & 0.295 \\
\bottomrule
\end{tabular}
}
\end{table*}

\subsection{Model Capacity Ablation}

We compare a one-layer linear mapping with a fixed MLP mapping, and report the SOTA baseline as a reference. This ablation is not intended to introduce another variant of the proposed method; instead, it tests whether a simple linear temporal mapping can already be competitive on some datasets and whether the additional nonlinear capacity of an MLP is always necessary. The results indicate that larger MLPs are not uniformly better. On Exchange and several ETT settings, the linear mapping remains highly competitive, while ETTm and high-dimensional datasets benefit more from nonlinear MLP capacity.

\begin{table*}[p]
\centering
\scriptsize
\setlength{\tabcolsep}{3pt}
\renewcommand{\arraystretch}{0.82}
\caption{Complete horizon-wise capacity ablation on ETT and Exchange datasets. Linear-STAIR and MLP-STAIR use fixed backbone families; SOTA Baseline denotes the best published baseline among the methods compared in the main table.}
\label{tab:appendix_capacity_full}
\resizebox{\textwidth}{!}{
\begin{tabular}{llcccccc}
\toprule
\multirow{2}{*}{Dataset} & \multirow{2}{*}{Horizon} & \multicolumn{2}{c}{Linear-STAIR} & \multicolumn{2}{c}{MLP-STAIR} & \multicolumn{2}{c}{SOTA Baseline} \\
\cmidrule(lr){3-4}\cmidrule(lr){5-6}\cmidrule(lr){7-8}
& & MSE & MAE & MSE & MAE & MSE & MAE \\
\midrule
\multirow{4}{*}{ETTh1} & 96 & 0.382 & 0.389 & 0.375 & 0.396 & 0.375 & 0.400 \\
 & 192 & 0.433 & 0.420 & 0.441 & 0.436 & 0.429 & 0.421 \\
 & 336 & 0.477 & 0.446 & 0.484 & 0.447 & 0.481 & 0.458 \\
 & 720 & 0.486 & 0.479 & 0.543 & 0.498 & 0.498 & 0.482 \\
\midrule
\multirow{4}{*}{ETTh2} & 96 & 0.289 & 0.342 & 0.295 & 0.344 & 0.289 & 0.341 \\
 & 192 & 0.371 & 0.393 & 0.371 & 0.392 & 0.372 & 0.392 \\
 & 336 & 0.411 & 0.427 & 0.422 & 0.433 & 0.386 & 0.414 \\
 & 720 & 0.418 & 0.450 & 0.432 & 0.450 & 0.412 & 0.434 \\
\midrule
\multirow{4}{*}{ETTm1} & 96 & 0.329 & 0.364 & 0.313 & 0.354 & 0.320 & 0.357 \\
 & 192 & 0.372 & 0.385 & 0.354 & 0.376 & 0.361 & 0.381 \\
 & 336 & 0.403 & 0.405 & 0.386 & 0.398 & 0.390 & 0.404 \\
 & 720 & 0.467 & 0.441 & 0.452 & 0.435 & 0.454 & 0.439 \\
\midrule
\multirow{4}{*}{ETTm2} & 96 & 0.181 & 0.273 & 0.169 & 0.250 & 0.175 & 0.258 \\
 & 192 & 0.242 & 0.311 & 0.234 & 0.294 & 0.237 & 0.299 \\
 & 336 & 0.305 & 0.354 & 0.290 & 0.331 & 0.298 & 0.340 \\
 & 720 & 0.398 & 0.406 & 0.389 & 0.390 & 0.391 & 0.396 \\
\midrule
\multirow{4}{*}{Exchange} & 96 & 0.080 & 0.201 & 0.086 & 0.204 & 0.086 & 0.205 \\
 & 192 & 0.161 & 0.289 & 0.191 & 0.309 & 0.176 & 0.299 \\
 & 336 & 0.300 & 0.398 & 0.348 & 0.426 & 0.301 & 0.397 \\
 & 720 & 0.719 & 0.636 & 0.848 & 0.694 & 0.839 & 0.691 \\
\bottomrule
\end{tabular}
}
\end{table*}

\begin{table*}[p]
\centering
\scriptsize
\setlength{\tabcolsep}{3pt}
\renewcommand{\arraystretch}{0.82}
\caption{Complete horizon-wise capacity ablation on Weather and high-dimensional datasets.}
\label{tab:appendix_capacity_full_cont}
\resizebox{\textwidth}{!}{
\begin{tabular}{llcccccc}
\toprule
\multirow{2}{*}{Dataset} & \multirow{2}{*}{Horizon} & \multicolumn{2}{c}{Linear-STAIR} & \multicolumn{2}{c}{MLP-STAIR} & \multicolumn{2}{c}{SOTA Baseline} \\
\cmidrule(lr){3-4}\cmidrule(lr){5-6}\cmidrule(lr){7-8}
& & MSE & MAE & MSE & MAE & MSE & MAE \\
\midrule
\multirow{4}{*}{Weather} & 96 & 0.161 & 0.213 & 0.163 & 0.208 & 0.158 & 0.209 \\
 & 192 & 0.204 & 0.256 & 0.209 & 0.249 & 0.206 & 0.250 \\
 & 336 & 0.253 & 0.293 & 0.266 & 0.292 & 0.251 & 0.287 \\
 & 720 & 0.327 & 0.344 & 0.346 & 0.343 & 0.339 & 0.341 \\
\midrule
\multirow{4}{*}{Traffic} & 96 & -- & -- & 0.438 & 0.287 & 0.395 & 0.268 \\
 & 192 & -- & -- & 0.452 & 0.296 & 0.417 & 0.276 \\
 & 336 & -- & -- & 0.470 & 0.305 & 0.433 & 0.283 \\
 & 720 & -- & -- & 0.502 & 0.323 & 0.467 & 0.302 \\
\midrule
\multirow{4}{*}{Electricity} & 96 & -- & -- & 0.148 & 0.245 & 0.148 & 0.240 \\
 & 192 & -- & -- & 0.162 & 0.259 & 0.162 & 0.253 \\
 & 336 & -- & -- & 0.177 & 0.276 & 0.178 & 0.269 \\
 & 720 & -- & -- & 0.208 & 0.306 & 0.220 & 0.310 \\
\midrule
\multirow{4}{*}{Solar} & 96 & -- & -- & 0.198 & 0.254 & 0.189 & 0.237 \\
 & 192 & -- & -- & 0.214 & 0.268 & 0.222 & 0.261 \\
 & 336 & -- & -- & 0.221 & 0.272 & 0.231 & 0.273 \\
 & 720 & -- & -- & 0.229 & 0.283 & 0.223 & 0.275 \\
\bottomrule
\end{tabular}
}
\end{table*}

\section{Limitations and Future Work}

This paper focuses on whether simple temporal mappings can be organized more effectively through a stagewise training paradigm. The proposed framework provides a clear way to distinguish shared temporal regularities, variable-wise adaptation, and cross-variable residual information, but it still has several limitations.

First, the current main results are reported with a single random seed. Although this practice is common in several LTSF baselines, multi-seed statistics would provide a more complete view of training stability. Second, the gains from Stage 2 and Stage 3 are dataset-dependent. This suggests that variable-specific information and cross-variable relationships are more suitable as residual sources rather than being assumed to be dominant in every dataset. Third, $\alpha$-RevIN uses a fixed normalization strength. Although it provides a simple way to analyze the role of local statistics, how to automatically select an appropriate normalization strength remains an open problem. Finally, the current cross-variable residual design is intentionally lightweight. More stable and better constrained residual modules may further improve performance on high-dimensional datasets.

Future work can explore adaptive but sufficiently regularized residual adapters, more robust low-rank cross-variable structures, and normalization strategies that preserve useful instance-level statistics while mitigating distribution shift.

\end{document}